\documentclass{article}

\usepackage{arxiv}

\usepackage[utf8]{inputenc} 
\usepackage[T1]{fontenc}    
\usepackage{hyperref}       
\usepackage{url}            
\usepackage{booktabs}       
\usepackage{amsfonts}       
\usepackage{nicefrac}       
\usepackage{microtype}      
\usepackage{xcolor}         
\usepackage{bm}
\usepackage{url}
\usepackage{amsmath}
\usepackage{amsthm}
\usepackage{amsfonts}
\usepackage{amssymb}
\usepackage{cancel}
\usepackage{graphicx}
\usepackage{natbib}
\usepackage{multirow}
\usepackage{makecell}
\usepackage{enumitem}
\usepackage{tabu}
\usepackage{wrapfig}

\usepackage{algorithm}
\usepackage{algorithmic}
\usepackage{subcaption}

\usepackage{hyperref}

\newtheorem{thm}{Theorem}
\newtheorem{prop}{Proposition}

\newtheorem{lemma}{Lemma}

\newtheorem{remark}{Remark}

\DeclareMathOperator*{\argmin}{arg\,min}

\newcommand{\Rttr}{R^2_{\text{\normalfont tr}}}
\newcommand{\Rtte}{R^2_{\text{\normalfont te}}}
\newcommand{\thetahv}{\bm{\hat{\theta}}}
\newcommand{\Kntk}{\bm{K}^{\text{\normalfont NTK}}}
\newcommand{\KntkP}{\bm{K}_{\bm{P}}^{\text{\normalfont NTK}}}
\newcommand{\Ph}{\bm{\Phi}}
\newcommand{\Pht}{\bm{\Phi_t}}

\newcommand{\muv}{\bm{\mu}}
\newcommand{\E}{\mathbb{E}}
\newcommand{\Var}{\text{\normalfont Var}}
\newcommand{\Cov}{\text{\normalfont Cov}}
\newcommand{\R}{\mathbb{R}}
\newcommand{\I}{\bm{I}}
\newcommand{\nv}{\bm{0}}
\newcommand{\Oo}{\mathcal{O}}

\newcommand{\dktwobar}{\overline{\Delta k_2}}
\newcommand{\dktwo}{\Delta k_2}
\newcommand{\dkv}{\bm{\Delta k}}
\newcommand{\smin}{s_{\min}}
\newcommand{\sminbar}{\overline{s_{\min}}}
\newcommand{\smax}{s_{\max}}

\newcommand{\ns}{n^*}
\newcommand{\vv}{\bm{v}}
\newcommand{\xv}{\bm{x}}
\newcommand{\xsv}{\bm{x^*}}
\newcommand{\xpv}{\bm{x'}}
\newcommand{\xvi}{\bm{x_i}}
\newcommand{\xvj}{\bm{x_j}}
\newcommand{\xsvi}{\bm{x^*_i}}
\newcommand{\X}{\bm{X}}
\newcommand{\Xs}{\bm{X^*}}
\newcommand{\yv}{\bm{y}}
\newcommand{\K}{\bm{K}}
\newcommand{\kv}{\bm{k}}

\newcommand{\Ks}{\bm{K^*}}
\newcommand{\Ss}{{\bm{\Sigma}}}

\newcommand{\fh}{\hat{f}}

\newcommand{\fv}{\bm{f}}
\newcommand{\fhv}{\bm{\hat{f}}}
\newcommand{\fhsv}{\bm{\hat{f}^*}}

\newcommand{\A}{\bm{A}}
\newcommand{\B}{\bm{B}}
\newcommand{\C}{\bm{C}}
\newcommand{\D}{\bm{D}}

\newcommand{\etahv}{\bm{\hat{\eta}}}
\newcommand{\df}{\text{\normalfont df}}
\newcommand{\Tr}{\text{\normalfont Tr}}

\title{Changing the Kernel During Training Leads to Double Descent in Kernel Regression}

\author{%
  Oskar Allerbo\\
  Department of Mathematics\\
  KTH Royal Institute of Technology\\
  \texttt{oallerbo@kth.se} \\
}

\begin{document}

\maketitle

\begin{abstract}
We investigate changing the bandwidth of a translational-invariant kernel during training when solving kernel regression with gradient descent. We present a theoretical bound on the out-of-sample generalization error that advocates for decreasing the bandwidth (and thus increasing the model complexity) during training. We further use the bound to show that kernel regression exhibits a double descent behavior when the model complexity is expressed as the minimum allowed bandwidth during training. Decreasing the bandwidth all the way to zero results in benign overfitting, and also circumvents the need for model selection. We demonstrate the double descent behavior on real and synthetic data and also demonstrate that kernel regression with a decreasing bandwidth outperforms that of a constant bandwidth, selected by cross-validation or marginal likelihood maximization.
We finally apply our findings to neural networks, demonstrating that by modifying the neural tangent kernel (NTK) during training, making the NTK behave as if its bandwidth were decreasing to zero, we can make the network overfit more benignly, and converge in fewer iterations.
\end{abstract}



\section{Introduction}
Double descent, the phenomenon that complex models tend to generalize poorly, while even more complex models often generalize well, has been observed and studied in numerous classes of models; for an overview see the \emph{Related Work} in Section \ref{sec:rel_work}. The phenomenon is closely related to benign overfitting, the observation that perfect interpolation of training data can be combined with good generalization.
Although these phenomena have been previously studied for kernel ridge regression, KRR, the focus has mainly been on benign overfitting, rather than on double descent, and predominantly in the high dimensional and/or asymptotic regimes. Additionally, in previous work, complexity is expressed through the dimensionality of the data, observing a peak in the generalization error when the number of features approximately equals the number of observations, i.e.\ when $n\approx d$. Thus, complexity is a function of the data, and not of the model.

In this paper, we look at model complexity, rather than data complexity, by expressing complexity through the chosen kernel. We study how kernel regression can be made to exhibit both benign overfitting and double descent when solved iteratively with gradient descent (kernel gradient descent, KGD) for a kernel whose complexity increases during training. 

\begin{figure}[t]
\center
\includegraphics[width=\textwidth]{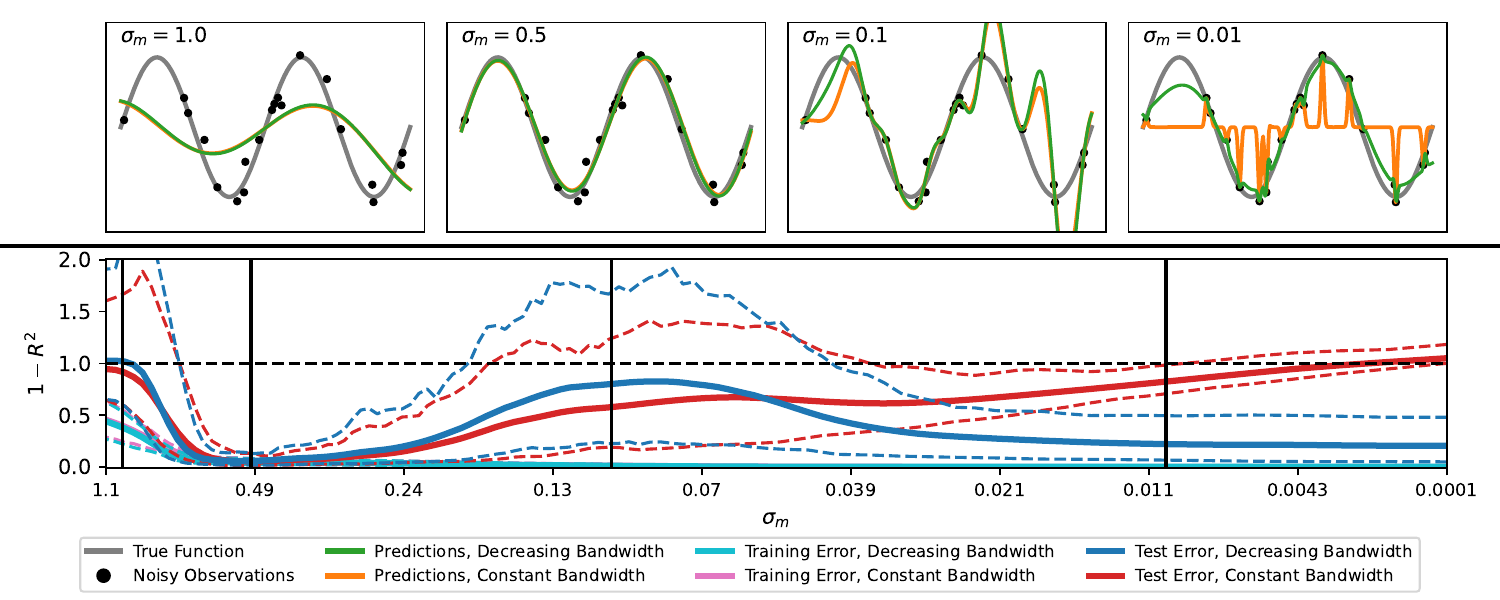}
\caption{Top: Inferred models for different values of the minimum allowed bandwidth, $\sigma_m$, and a fixed training time/regularization, see Appendix \ref{sec:exp_dets} for details. With increasing model complexity (decreasing $\sigma_m$) the generalization goes from poor, to good, to poor, to, for the decreasing bandwidth model, good. Bottom: Mean and 90\% prediction intervals of training and generalization errors (as $1-R^2$)\protect\footnotemark\ as a function of $\sigma_m$. The four vertical black lines correspond to the four top panels.}
\label{fig:syn_dd_100}
\end{figure}

Intuitively, and as discussed in detail in Appendix \ref{sec:bw_compl}, for KRR, and KGD, with a translational-invariant kernel, $k(\xv,\xpv,\sigma)=k(\|\xv-\xpv\|_2/\sigma)$, the complexity of the inferred function is governed by the kernel bandwidth, $\sigma\geq 0$, where a smaller bandwidth results in a more complex function and vice versa. Thus, for KGD, where it is possible to change the kernel in every iteration, decreasing the bandwidth during training, corresponds to gradually increasing the model complexity. This results in simpler relations in the data being captured earlier during training, while more complex relations are captured later. 
If the bandwidth goes to zero during training, the model can perfectly interpolate the training data and still generalize well, as exemplified in the top right panel of Figure \ref{fig:syn_dd_100}, it overfits benignly.

However, if, for a constant training time, the maximum complexity of the model is restricted through a minimum allowed bandwidth $\sigma_m$, we may observe a double descent behavior in $\sigma_m$, as demonstrated in Figure \ref{fig:syn_dd_100}. Here, we compare KGD with decreasing bandwidth, KGD-D, to KRR with constant bandwidth. For KRR, $\sigma_m$ was used during the entire training, and for KGD-D, it was used as the minimum allowed bandwidth.
For $\sigma_m=1$, the inferred models are too simple and perform badly on both training and test data. For $\sigma_m=0.5$, the models do well on both training and test data. For $\sigma_m=0.1$, the models almost perfectly interpolate the training data but at the expense of extreme predictions between observations, and generalize poorly. For $\sigma_m=0.01$, since simpler models are used in the early stages of training, KGD-D overfits benignly, i.e.\ it perfectly explains the training data and also performs well on test data. 
In contrast, the constant bandwidth model, where $\sigma_m=0.01$ is used during the entire training, is complex enough to incorporate the training data and still predict the prior, the zero function, between observations, which results in poor generalization.

\footnotetext{$R^2:=1-\|\yv-\fhv\|_2^2/\|\yv\|_2^2\leq 1$, measures the proportion of the variance in $\yv$ that is explained by the model. $R^2=1$ corresponds to a perfect fit, while a model that predicts $\fhv=\nv$ results in $R^2=0$; thus a negative value of $R^2$ is possible, but it corresponds to a model that performs worse than always predicting zero.}

Our \textbf{main contributions} are:
\begin{itemize}[leftmargin=6mm]
\item
We derive a \emph{generalization bound} for KGD-D, which we use both as a \emph{motivation for decreasing the bandwidth} during training, and for \emph{explaining the observed double descent} phenomenon of KGD-D as a function of minimum allowed bandwidth.
\item
We propose a \emph{scheme for how to decrease the bandwidth} of KGD-D during training and show on real data that with this scheme, \emph{KGD-D outperforms KRR with a constant bandwidth}, selected by cross-validation or marginal likelihood maximization, in terms of model performance.
\item
We transfer our insights from KGD-D to \emph{neural networks}: By \emph{actively modifying the neural tangent kernel} during training, to mimic a decreasing bandwidth, we can both make the network \emph{overfit more benignly} and \emph{converge in fewer iterations}.
\end{itemize}

All proofs are deferred to Appendix \ref{sec:proofs}; code is available at \url{https://github.com/allerbo/non_constant_kgd}.

\section{Related Work}
\label{sec:rel_work}
\textbf{Kernel Ridge Regression}: KRR is a generalization of linear ridge regression that is non-linear in the data, but linear in the parameters. The KRR estimate coincides with the posterior mean of kriging, or Gaussian process regression \citep{krige1951statistical,matheron1963principles} and has successfully been applied within a wide range of applications, including
geoscience \citep{fan2021well,safari2021kernel},
hydrology \citep{ali2020complete},
ecology \citep{singh2021neural},
genomics \citep{yao2025genomic, liu2025investigation},
chemistry \citep{zahrt2019prediction, shahsavar2021experimental},
physics \citep{wu2021increasing, bortolussi2025towards},
manufacturing \citep{chen2021optimizing},
signal and image processing \citep{le2021fingerprinting, chen2022kernel}, and
transportation \citep{he2025rtsmffde}. 
Solving kernel regression with gradient descent, something we refer to as kernel gradient descent, KGD, but which is also known as functional gradient descent \citep{mason1999boosting}, has been studied by e.g.\ \citet{yao2007early}, \citet{raskutti2014early}, \citet{ma2017diving}, and \citet{allerbo2025fast}. 

\textbf{Benign overfitting} and \textbf{double descent}: Early work on the topic include those by \citet{bartlett2002rademacher}, \citet{zhang2016understanding}, \citet{dziugaite2017computing}, \citet{belkin2018overfitting, belkin2019reconciling}, and \citet{advani2020high}. The subject has been studied for many classes of models, including linear and logistic regression \citep{bartlett2020benign, hastie2022surprises, mei2022generalization, deng2022model}, k-nearest neighbors \citep{curth2024classical},
regression trees and boosting \citep{belkin2019reconciling, curth2024u}, principal component regression \citep{gedon2024no}, neural networks \citep{geiger2020scaling, nakkiran2021deep, ju2021generalization},
and kernel regression \citep{poggio2019double, liu2021kernel}. 
For kernel regression, double descent has previously been studied as a function of data complexity, in contrast to our work, which studies double descent as a function of model complexity. 
For KRR, benign overfitting has been studied by \citet{liang2020just}, \citet{buchholz2022kernel} and \citet{li2024kernel}, \citet{mallinar2022benign}, \citet{barzilai2023generalization}, and \citet{cheng24characterizing}, where the latter three have studied the fact that benign overfitting cannot occur unless the decay of the kernel eigenvalues is slow. \citet{haas2023mind} obtained benign overfitting by using a constant kernel that is a mixture of a high and a low bandwidth kernel.

\textbf{The neural tangent kernel}: The connection between neural networks and Gaussian processes was established by \citet{neal1995bayesian} and has since been studied by e.g.\ \citet{lee2017deep}, \citet{matthews2018gaussian}, \citet{garriga2018deep}, and \citet{yang2019wide}.
\citet{jacot2018neural} showed that in the infinite-width limit, training a fully connected neural network with gradient descent is identical to KGD with a kernel that depends on the initialization of the network parameters. The authors refer to this kernel as the neural tangent kernel, NTK. The NTK has since become a popular tool for analyzing neural networks \citep{bietti2019inductive, geifman2020similarity, liu2020linearity, adlam2020neural, tsilivis2022can}, and has been extended to several different network architectures, including convolutional \citep{arora2019exact}, graph \citep{du2019graph} and recurrent \citep{alemohammad2022recurrent} neural networks.
While for infinitely wide networks, the NTK is constant during training, this is not the case for neural networks with finitely many parameters, whose complexities are known to increase with training \citep{kalimeris2019sgd, loo2022evolution}, where the latter specifically showed how the complexity of the NTK increases during training. 
\cite{geifman2023controlling} modified the NTK by preconditioned gradient descent to increase the training speed, however, in contrast to our work, without addressing the implications on generalization.

\section{Kernel Gradient Descent with Decreasing Bandwidth}
\label{sec:kgdd}

\subsection{Notation and Assumptions}
We let $\X=[\bm{x_1},\bm{x_2},\dots \bm{x_n}]^\top\in\R^{n\times d}$ denote the in-sample (training) design matrix, $\yv \in \R^n$ the response vector, and $\Xs=[\bm{x^*_1},\bm{x^*_2},\dots \bm{x^*_{n^*}}]^\top\in\R^{\ns\times d}$ the out-of-sample design matrix. We assume without loss of generality that the data is centered, i.e.\ that $\sum_{i=1}^ny_i=0$. If this is not the case, the data is centered by subtracting the mean of $\yv$ before training and adding it back after training.
For a positive semi-definite, PSD, kernel function, $k(\xv,\xpv)\geq 0$, the two kernel matrices $\K=\K(\X)\in \R^{n\times n}$ and $\Ks=\K(\Xs,\X)\in\R^{\ns\times n}$ 
are defined according to $\K_{ij}=k(\xvi,\xvj)$ and $\Ks_{ij}=k(\xsvi,\xvj)$, while $\fhv=\fhv(\X)=[\fh(\bm{x_1}),\fh(\bm{x_2}),\dots \fh(\bm{x_n})]^\top\in \R^{n}$ and $\fhsv=\fhv(\Xs,\X)=[\fh(\bm{x^*_1}),\fh(\bm{x^*_2}),\dots \fh(\bm{x^*_{\ns}})]^\top\in \R^{\ns}$ denote in- and out-of-sample model predictions.
For a PSD matrix $\A$, the weighted norm of vector $\vv$ is defined as $\|\vv\|_{\A}:=\sqrt{\vv^\top\A\vv}$.

\subsection{Introducing KGD-D}
The update rule for KGD, with learning rate $\eta$, is given by
\begin{equation}
\label{eq:kgd_update}
\begin{bmatrix}\fhv_{k+1}\\\fhsv_{k+1}\end{bmatrix}=\begin{bmatrix}\fhv_k\\\fhsv_k\end{bmatrix}+\eta\cdot\begin{bmatrix}\K_k\\\Ks_k\end{bmatrix}\left(\yv-\fhv_{k}\right).
\end{equation}
By treating Equation \ref{eq:kgd_update} as an Euler forward formulation, the corresponding differential equation for kernel gradient flow, KGF, is obtained as
\begin{equation}
\label{eq:kgd_diff_eq}
\frac{\partial}{\partial t}\left(\begin{bmatrix}\fhv(t)\\\fhsv(t)\end{bmatrix}\right)=\begin{bmatrix}\K(t)\\\Ks(t)\end{bmatrix}\left(\yv-\fhv(t)\right),
\end{equation}
which, when $\K$ and $\Ks$ are constant with respect to $t$, and $[\fhv(0)^\top,\fhsv(0)^\top]^\top=\nv$, has the closed form solution
\begin{equation}
\label{eq:kgf_s}
\begin{bmatrix}\fhv(t)\\\fhsv(t)\end{bmatrix}=\begin{bmatrix}\I\\\Ks\K^{-1}\end{bmatrix}(\I-\exp(-t\K))\yv,
\end{equation}
where $\exp$ denotes the matrix exponential. This can be compared to the KRR solution,
\begin{equation}
\label{eq:krr_s}
\begin{aligned}
\begin{bmatrix}\fhv(\lambda)\\\fhsv(\lambda)\end{bmatrix}
=\begin{bmatrix}\K\\\Ks\end{bmatrix}(\K+\lambda\cdot\I)^{-1}\yv
=\begin{bmatrix}\I\\\Ks\K^{-1}\end{bmatrix}(\I-(\I+1/\lambda\cdot\K)^{-1})\yv.
\end{aligned}
\end{equation}
\begin{remark}
Since $\exp(-t\K)=\exp(t\K)^{-1}$, KRR can be thought of as a first-order Taylor approximation of KGF for $\lambda=1/t$.
\end{remark}

When the kernel, and thus $\K$ and $\Ks$, is allowed to change with $t$, Equation \ref{eq:kgd_diff_eq} no longer has a closed-form solution. We can, however, use the relation $\partial_t \fh(\xsv,t)=\kv(\xsv,\X,t)^\top(\yv-\fhv(\X,t))$ from Equation \ref{eq:kgd_diff_eq}, where $\kv(\xsv,\X,t)\in\R^n$ is the row in $\Ks$ that corresponds to $\xsv$, to bound the out-of-sample generalization error of the KGF estimate, as is done in Theorem \ref{thm:gen_bound}.

\subsection{Generalization Properties of KGD-D}
\begin{thm}~\\
\label{thm:gen_bound}
Assume that the true function, $f$, has bounded derivatives and that $y_i=f(\xvi)+\varepsilon_i$, where $\E|\varepsilon_i|<\infty$ for all $i$.

Let $\smin$ denote the smallest singular value of $\K$, $\smin(\X,t):=\smin(\K(\X,t))$, and let $\dktwo(\xsv,\X,t):= \sqrt{n} \cdot \|\kv(\xsv,\X,t)-\frac1n\sum_{i=1}^n\kv(\xvi,\X,t)\|_2$, quantify the difference between the out-of- and in-sample kernel evaluations.

Let $\sminbar$ and $\dktwobar$ denote their averages during training, for $\dktwobar$ with weight function $\|\yv-\fhv(\X,t)\|_2$:
\begin{equation*}
\begin{aligned}
&\sminbar(\X,t):=\frac1t\int_0^t\smin(\X,\tau)d\tau,\ \dktwobar(\xsv,\X,t):=\frac{\int_0^t\dktwo(\xsv,\X,\tau)\left\|\yv-\fhv\left(\X,\tau\right)\right\|_2d\tau}{\int_0^t\left\|\yv-\fhv\left(\X,\tau\right)\right\|_2d\tau}.
\end{aligned}
\end{equation*}

Then, the expected out-of-sample generalization error is bounded by
\begin{equation}
\label{eq:gen_bound}
\begin{aligned}
&\E\left(\left|\fh(\xsv,\X,t)-f(\xsv)\right|\right)
\leq\left(e^{-t\cdot\sminbar(\X,t)} + \dktwobar(\xsv,\X,t)\cdot\int_0^t e^{-\tau\cdot\sminbar(\X,\tau)}d\tau \right)\cdot\frac{\|\yv\|_2}{\sqrt n}+C,
\end{aligned}
\end{equation}
where the expectation is with respect to $\bm{\varepsilon}$, and where $C$ depends on $f$, $\X$, $\xsv$, and $\E|\varepsilon|$, but not on the kernel or the training time, $t$.
\end{thm}
\begin{remark}Since $\K$ is a PSD matrix at all times, its singular values coincide with its eigenvalues.
\end{remark}
Let us discuss the two contributions to the generalization bound in Equation \ref{eq:gen_bound} a bit closer.
\begin{itemize}[leftmargin=.5cm]
\item
The first term, $e^{-t\cdot\sminbar}\cdot\|\yv\|_2^2/\sqrt{n}$, that does not depend on $\xsv$, is connected to the training error and eventually goes to zero during training. However, if $\sminbar$ is small, this process might be slow. This is why generalization is poor in the first panel in Figure \ref{fig:syn_dd_100}.
\item
The second term depends on $\xsv$ through $\dktwo$, and captures the difference of the inferred function on training data and new data (in- and out-of-sample). If this difference is large, generalization may be poor, despite good performance on training data. This is what happens in the third and fourth panels in Figure \ref{fig:syn_dd_100} (in the fourth panel only for constant bandwidth).
\end{itemize}
To better understand when $\dktwo$ is large, we treat it as a random function in $\xv$. In Proposition \ref{thm:dktwo_bd}, we bound its expected value for translational invariant kernels.
\begin{prop}~\\
\label{thm:dktwo_bd}
Assume that $k(\xv,\xpv)=k\left(\|\xv-\xpv\|_2\right)$, that $\xsv,\bm{x_1},\dots \bm{x_n}$ are iid random variables and denote $\mu_k:=\E(k(\xv,\xpv))$, $\varsigma_k^2:=\Var(k(\xv,\xpv))$, and $c_k:=\Cov(k(\xv,\xpv),k(\xv,\bm{x''}))$, where $\xv\neq \xpv$, $\xv\neq \bm{x''}$, $\xpv\neq \bm{x''}$.


Then
\begin{equation}
\label{eq:dktwo_bd}
\begin{aligned}
\left(\E\left(\dktwo(\xsv,\X)\right)\right)^2
&\leq\E\left(\dktwo(\xsv,\X)^2\right)
=(\mu_k-k(0))^2+(n^2+n-2)\cdot(\varsigma_k^2-c_k)+\varsigma^2_k\geq 0.
\end{aligned}
\end{equation}
\end{prop}
From Equation \ref{eq:dktwo_bd}, we see that $\dktwo$ is large when the variance of $k$ is large. Note, however, that for a translational-invariant kernel of the form $k(\xv,\xpv,\sigma)=k\left(\|\xv-\xpv\|_2/\sigma\right)$, $k$ is independent of the data for $\sigma=0$ and $\sigma=\infty$: $k(\|\xv-\xpv\|_2/0)=k(\infty)$ and $k(\|\xv-\xpv\|/\infty)=k(0)$. In these cases, there is no randomness, and, for kernels with $k(0)=1$ and $k(\infty)=0$, we obtain
\begin{equation*}
\begin{aligned}
&\dktwo(\sigma=0)^2=\left(\mu_k(\sigma=0)-k(0)\right)^2=\left(k(\sigma=0)-k(0)\right)^2=(0-1)^2=1\\
&\dktwo(\sigma=\infty)^2=\left(\mu_k(\sigma=\infty)-k(0)\right)^2=\left(k(\sigma=\infty)-k(0)\right)^2
=(1-1)^2=0.
\end{aligned}
\end{equation*}

\begin{figure}
\center
\includegraphics[width=\textwidth]{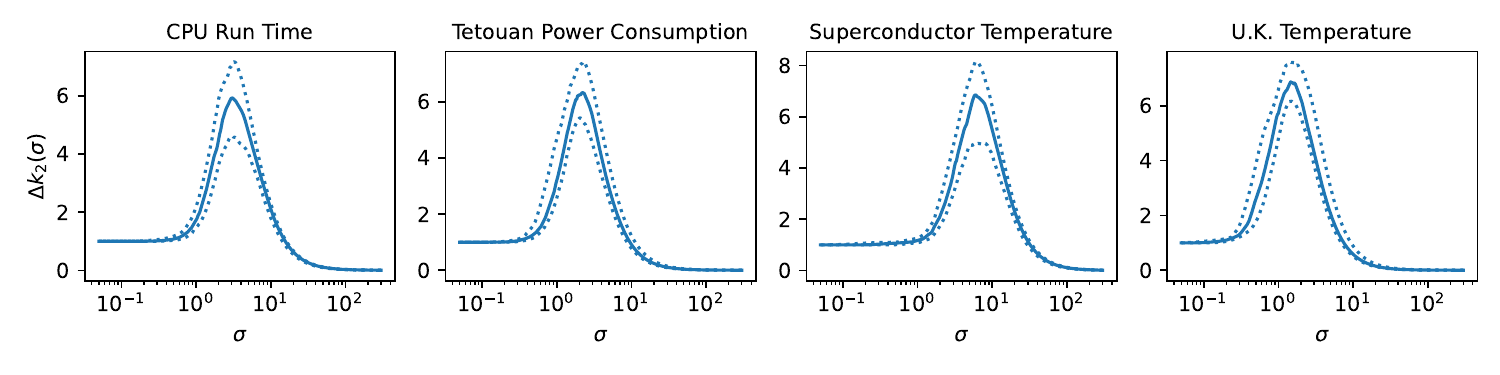}
\caption{$\dktwo$ as a function of $\sigma$. We used the Gaussian kernel, with the data standardized to zero mean and unit variance, and split into 100 random, non-overlapping splits, each containing 30 observations. The solid and dotted lines mark the median and first and fourth quartile across the 100 splits. As $\sigma$ goes to $0$ and $\infty$, $\dktwo$ becomes deterministically 1 and 0, while for intermediate values of $\sigma$, $\dktwo$ exhibits randomness and is significantly larger than 1.}
\label{fig:k2_demo}
\end{figure}

For intermediate values of $\sigma$, $\varsigma^2_k\neq 0$, and $\E\left(\dktwo(\xsv,\X,\sigma)^2\right)=n^2\cdot(\varsigma_k^2-c_k)+\Oo(n)$, where $\Oo(n)$ denotes terms of order $n$ or lower. 
In Figure \ref{fig:k2_demo}, we exemplify this on the four real data sets in Table \ref{tab:datasets}. As expected from Proposition \ref{thm:dktwo_bd}, $\dktwo\approx1$ when $\sigma$ is small, and $\dktwo\approx 0$ when $\sigma$ is large, with a, random, maximum in between. We refer to the three regions as the one-region, the peak-region and the zero-region.

%
Let us now address the implications for generalization. If $\dktwo$ is kept in the zero-region, i.e.\ if $\sigma$ is large, during the entire training, $\dktwobar$ will be small, which is needed for good generalization. However, a large bandwidth results in $\K$ being close to singular, and thus in poor generalization due to a small $\sminbar$. Thus, if $\sigma$ were to be held constant during the training, a good compromise value had to be chosen. If we, however, allow $\sigma$ to change during training, we can use the fact that $\dktwobar$ is weighted by $\|\yv-\fhv(\X,t)\|_2$, a function that decreases during training, while $\sminbar$ is not: We obtain a small $\dktwobar$ if $\sigma$ is large during the early stages of training, and we avoid a small $\sminbar$ if $\sigma$ is small during \emph{some} part of the training.

This suggests that generalization will be good if we start with a large bandwidth, to avoid a too large $\dktwobar$, which we gradually decrease toward zero during training, to avoid a too small $\sminbar$. This agrees with the conclusions above, and also has an interesting connection to gradient boosting, as discussed in Appendix \ref{sec:fsam}. 
There is, however, still a risk that generalization will be poor due to the bandwidth decreasing too fast, in which case $\dktwobar$ will be too large, or too slowly, in which case $\sminbar$ will be too small.
In Section \ref{sec:bw_dec}, we propose a simple method for selecting the bandwidth decreasing speed.

\begin{wrapfigure}{R}{9.2 cm}
\vspace{-.8cm}
\begin{minipage}{9.2 cm}
\begin{algorithm}[H]
\caption{Kernel Gradient Descent with Decreasing Bandwidth, KGD-D.}
\textbf{Input:} Training data, $(\X,\yv)$. Out-of-sample covariates, $\Xs$. Initial bandwidth, $\sigma_0$. Minimum allowed bandwidth, $\sigma_m$. Learning rate, $\Delta t$. Minimum $\Rttr$ speed, $v_{R^2}$. Maximum $\Rttr$, $R^2_{\text{tr},\max}$.\\
\textbf{Output:} Prediction vector, $[\fhv(t)^\top, \fhsv(t)^\top]^\top$.\\
\begin{algorithmic}[1]
  \STATE Initialize $[\fhv(0)^\top, \fhsv(0)^\top]^\top=\nv$.
  \STATE Initialize $\sigma(0)=\sigma_0$, $\K(0)=\K(\sigma_0)$\\ and $\Ks(0)=\Ks(\sigma_0)$.
  \REPEAT
  \STATE Calculate $\Rttr(t)=1-\frac{\|\yv-\fhv(t)\|_2^2}{\|\yv\|_2^2}$\\ and $\frac{\partial \Rttr(t)}{\partial t}=2\cdot\frac{\|\yv-\fhv(t)\|_{\K(t)}^2}{\|\yv\|_2^2}$.
  \IF {$\frac{\partial \Rttr(t)}{\partial t}< v_{R^2}$}
    \REPEAT
     \STATE Set $\sigma(t)=\frac{\sigma(t)}{1+\Delta t}$, and calculate $\K(\sigma(t))$ and $\frac{\partial \Rttr(t)}{\partial t}$
    \UNTIL {$\frac{\partial \Rttr(t)}{\partial t}\geq v_{R^2}$ \textbf{or} $\sigma(t)\leq \sigma_m$}.
    \STATE Calculate $\Ks(\sigma(t))$.
  \ENDIF
  \STATE Update $[\fhv(t)^\top, \fhsv(t)^\top]^\top$ according to
  \begin{equation*}
  \begin{aligned}
  &\begin{bmatrix} \fhv(t+\Delta t)\\ \fhsv(t+\Delta t)\end{bmatrix}
  =\begin{bmatrix} \fhv(t)\\ \fhsv(t)\end{bmatrix}
  +\Delta t \begin{bmatrix} \K(t)\\ \Ks(t)\end{bmatrix}\left(\yv-\fhv(t)\right).
  \end{aligned}
  \end{equation*}
  \STATE Set $t=t+\Delta t$.
  \UNTIL{$\Rttr(t)\geq R^2_{\text{tr},\max}$}.
\end{algorithmic}
\label{alg:kgdd}
\end{algorithm}
\end{minipage}
\vspace{-1.5cm}
\end{wrapfigure}

\subsection{Double Descent in Minimum Bandwidth}
\label{sec:dd_kgdd}
We will now use Theorem \ref{thm:gen_bound} to explain the double descent behavior observed in Figure \ref{fig:syn_dd_100}. 
While $\sminbar$ increases with increasing model complexity (decreasing $\sigma_m$), $\dktwobar$ first increases, and then decreases as $\sigma_m$ decreases.
By combining these two behaviors, we can obtain the following four phases, which explain the behavior in Figure \ref{fig:syn_dd_100}:
\begin{itemize}[leftmargin=.5cm]
\item For \textbf{low model complexity} (large $\sigma_m$), $\dktwobar$ is small, since $\dktwo$ is the zero region during the entire training, but $e^{-t\sminbar}$ is large, since $\sminbar$ is small, and generalization is bad. This happens at $\sigma_m=1.0$.
\item For \textbf{moderate model complexity} (moderate $\sigma_m$), $\dktwobar$ and $e^{-t\sminbar}$ are both small, and generalization is good. This happens at $\sigma_m=0.5$.
\item For \textbf{high model complexity} (small $\sigma_m$), $e^{-t\sminbar}$ is small, but $\dktwobar$ is large, since too much training is done when $\dktwo$ is in the peak-region. This happens at $\sigma_m=0.1$.
\item For \textbf{very high model complexity} (very small $\sigma_m$), both $e^{-t\sminbar}$ and $\dktwobar$ are small, where the latter is a consequence of $\dktwo$ being almost only in in the zero- and one-regions during training. This happens at $\sigma_m=0.01$.
\end{itemize}
Note that if a very small $\sigma$ is used during the entire training, as is done for KRR with constant bandwidth in the top right panel of the figure, $\dktwo$ is in the one-region during the entire training, and never in the zero region, which explains the bad generalization in this case.

\newpage
\subsection{The KGD-D Algorithm}
\label{sec:bw_dec}
The speed of the bandwidth decrease is crucial, since a too fast decay results in a too large $\dktwobar$, while a too slow decay results in too small $\sminbar$. We propose a bandwidth-decreasing scheme based on $R^2$ on training data, $\Rttr(t):=1-\|\yv-\fhv(\X,t)\|_2^2/\|\yv\|_2^2$.
According to Equation \ref{eq:dr2dt1} in Proposition \ref{thm:dr2dt}, for KGF, $\Rttr$ always increases during training, regardless of how the bandwidth is updated. However, if the kernel is kept constant, according to Equation \ref{eq:dr2dt3}, the speed of the increase decreases with training time. This means that, unless the kernel is updated, eventually the improvement of $\Rttr$, although always positive, will be very small. Based on this observation we use the following simple update rule:
\begin{itemize}[leftmargin=.5cm]
\item
If $\frac{\partial \Rttr(t)}{\partial t}=2\cdot\frac{\|\yv-\fhv(\X,t)\|^2_{\K(t)}}{\|\yv\|_2^2}<v_{R^2}$, decrease the bandwidth.
\end{itemize}
That is if the increase in $\Rttr$ is smaller than some threshold value, $v_{R^2}$, then the bandwidth is decreased until the increase is again fast enough. Otherwise, the bandwidth is kept constant. The exact step size of the decrease is less important, as long as it is not too large. We use $\sigma=\sigma/(1+\Delta t)$, where $\Delta t$ is the learning rate.

\begin{prop}~\\
\label{thm:dr2dt}
Let $R^2$ on training data be defined according to $\Rttr(t):=1-\|\yv-\fhv(\X,t)\|_2^2/\|\yv\|_2^2$, where $\fhv(\X,t)$ is the KGF estimate.
Then, for a, possibly non-constant, kernel $\K(t)$,
\begin{equation}
\label{eq:dr2dt1}
\frac{\partial \Rttr(t)}{\partial t}=2\cdot\frac{\|\yv-\fhv(\X,t)\|^2_{\K(t)}}{\left\|\yv\right\|_2^2}\geq 0.
\end{equation}
For a constant kernel $\K$,
\begin{equation}
\label{eq:dr2dt3}
\frac{\partial^2 \Rttr(t)}{\partial t^2}=-4\cdot\frac{\|\yv-\fhv(\X,t)\|^2_{\K^2}}{\left\|\yv\right\|_2^2}\leq 0.
\end{equation}
\end{prop}

KGD-D is summarized in Algorithm \ref{alg:kgdd}.

For KRR (and KGD/KGF) with constant bandwidth, the hyperparameters $\lambda$ (or $t$) and $\sigma$ have to be carefully selected to obtain a good performance. This is usually done by cross-validation, where the data is split into training and validation data, or by marginal likelihood maximization, where the hyperparameters are optimized together with the model parameters. When starting with a large bandwidth that gradually decreases to zero during training and training until convergence, the issue of hyperparameter selection is in some sense circumvented since many different values of $\sigma$ and $t$ are used during the training. However, instead of bandwidth and regularization, the minimum $\Rttr$ speed, $v_{R^2}$, which controls the decrease of $\sigma$, has to be selected. But since $\Rttr$ is a normalized quantity, this hyperparameter tends to generalize well across data sets. As demonstrated in Section \ref{sec:exps}, $v_{R^2}=0.1$ tends to lead to good performance.

Just like for all kernel methods, scaling with data might pose an issue for KGD-D. While KRR with cross-validation scales as $n^3\cdot \#h$, where $\#h$ denotes the number of candidate hyper-parameters, KGD-D scales as $n^2\cdot \#i$, where $\#i$ denotes the number of iterations. Another limitation of KGD-D is that, as for all functional gradient descent methods, $\fhsv$ and $\fhv$ are calculated together, and thus $\Xs$ must be included during training.

\begin{figure}[b]
\center
\includegraphics[width=\textwidth]{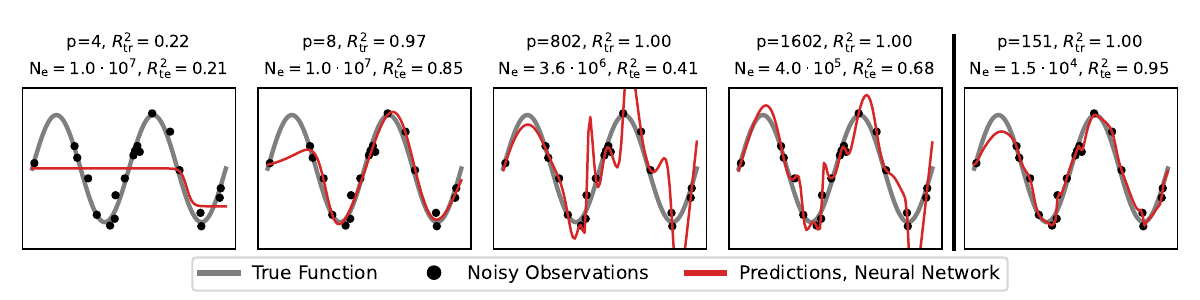}
\caption{Inferred functions of neural networks of increasing complexity, trained until $\Rttr$ exceeds 0.999, or at most for $10^7$ epochs, N$_\text{\normalfont e}$, see Appendix \ref{sec:exp_dets} for details. $\Rtte$ denotes $R^2$ on test data. In the fifth panel, we have used preconditioned gradient descent to make the NTK evolve as if its bandwidth decreased to zero during training.}
\label{fig:syn_nn}
\end{figure}

\section{Connection Between KGD-D and Neural Networks}
\label{sec:nn}
In the NTK framework, training a neural network amounts to KGD with the non-constant neural tangent kernel, which suggests that there is a strong connection between neural networks and KGD-D. This is supported by the fact that neural networks are known for exhibiting a double descent behavior, although with model complexity measured by the number of parameters rather than by kernel bandwidth.

In the first four panels of Figure \ref{fig:syn_nn}, we display the inferred functions of feed-forward neural networks of increasing complexity, in terms of the number of parameters, $p$, trained on the same data as in Figure \ref{fig:syn_dd_100}. As expected, the functions are similar to those inferred by KGD-D; however, in the fourth panel, even if the generalization is better than in the third panel, there is still room for improvement especially compared to KGD-D.

In the fifth panel, we trained the model with preconditioned gradient descent with a preconditioning matrix that allows us to change the NTK during training. With inspiration from KGD-D, we have made the NTK behave as if its bandwidth decreased to zero during training, see Appendix \ref{sec:ntk} for details. Even if these results are preliminary, the fact that the preconditioned model generalizes better, and converges in fewer iterations, suggests that KGD-D inspired training of neural networks may be an interesting line of future research.

\section{Experiments on Real Data}
\label{sec:exps}

\begin{table}
\center
\caption{Real data sets.}
\begin{tabular}{@{}l l@{}}
\toprule
Data Set (DS) & Size, $n\times d$\\
\midrule
Run time of CPUs \citep{delve1996comp} & $8192\times21$\\
Power consumption of Tetouan City \citep{salam2018comparison} & $52416\times7$\\
Critical temperature of superconductors \citep{hamidieh2018data} & $21263\times  81$\\
Daily temperature in the U.K.\ in the year 2000 \citep{wood2017generalized} & $45568\times 4$\\
\bottomrule
\end{tabular}
\label{tab:datasets}
\end{table}

\begin{figure*}[t]
\center
\includegraphics[width=\textwidth]{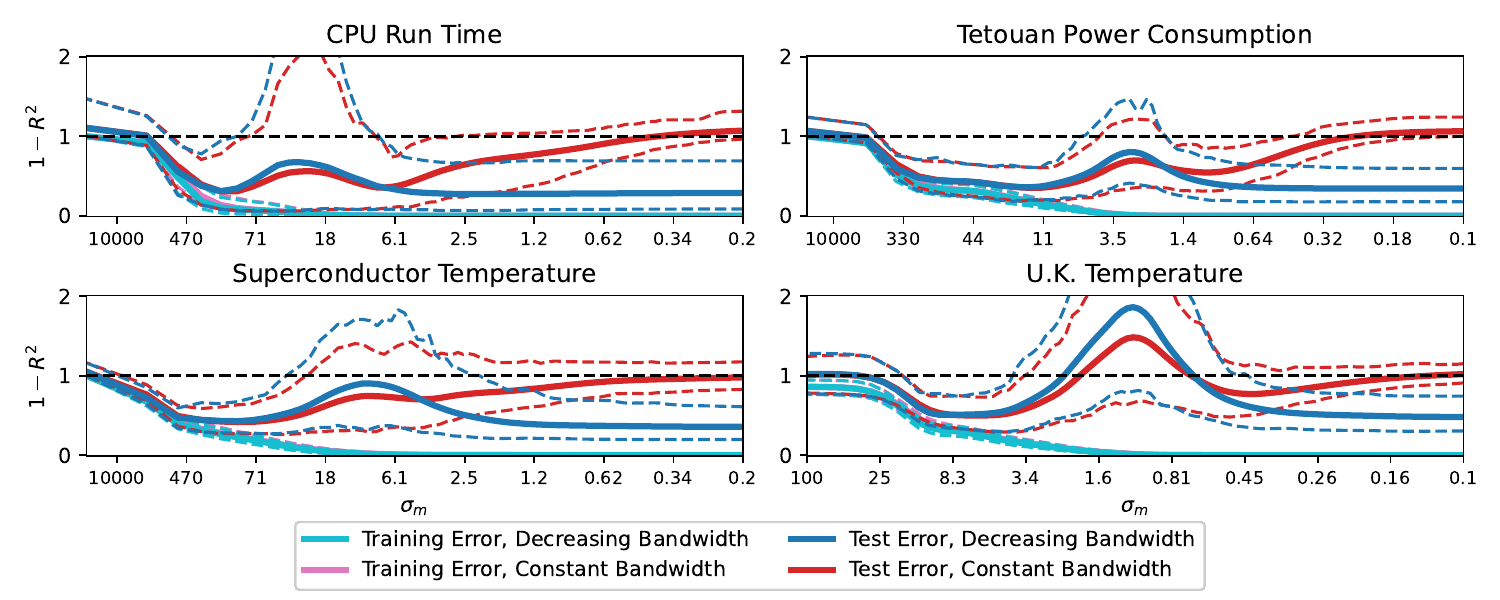}
\caption{Training and test errors (as $1-R^2$) as functions of model complexity (in terms of $\sigma_m$), for kernel regression with constant and decreasing bandwidths. The plots show the means, together with the 90\% prediction intervals. For simple models, both the training and test errors are large, but they all decrease with increasing model complexity. While the training errors decrease toward zero, the test errors start to increase again as the models become even more complex. When using a decreasing bandwidth, a second descent in the error results in good generalization for very complex models, something that is not the case for the constant bandwidth model.}
\label{fig:dd_100}
\end{figure*}

In this section, we demonstrate KGD-D, and compare it to KRR with constant bandwidth, on the four real data sets in Table \ref{tab:datasets}. The exact experimental setups are given in Appendix \ref{sec:exp_dets}; here follows a summary: The data were standardized to zero mean and unit variance and split into 100 (for Figure \ref{fig:dd_100}) or 200 (for Table \ref{tab:dec_100}) random, non-overlapping splits, each containing 100 observations (for the smaller CPU data set, we always used 100 splits, 93 of them of size 82 and 7 of size 81), which in turn were split 80\%/20\% into training and test data.
We consistently used the Gaussian kernel, $k(\xv,\xpv)=\exp\left(-\frac{\|\xv-\xpv\|^2_2}{2\sigma^2}\right)$; in Appendix \ref{sec:more_exps}, we also present results for four additional kernels, a neural network, and KGF.

In Figure \ref{fig:dd_100}, we demonstrate double descent as a function of model complexity, in terms of the minimum allowed bandwidth, $\sigma_m$, on the four datasets,
for a fixed regularization, $\lambda=1/t=0.001$. For KRR, $\sigma_m$ was used during the entire training, and for KGD-D, it was used as the minimum allowed bandwidth.

Just as for the synthetic data, for large values of $\sigma_m$, both models perform poorly both in terms of training and test error. With decreasing $\sigma_m$ (increasing model complexity), both the training and test errors decrease, but while the training errors always decrease, the test errors start to increase again. Approximately where the training errors become zero, there is a peak in the test errors, but while the test error for KGD-D continues to decrease with model complexity, the test error for KRR goes to $1-R^2=1$, i.e.\ $R^2=0$, as $\sigma_m$ goes to zero.

\begin{table}
\caption{Median and first and third quartile, over the 200 (or 100 for the CPU data) splits, of $R^2$ on the test data. The p-value is that of a Wilcoxon signed-rank test, testing whether the method performs better than KRR-CV and KRR-MML, where p-values less than 0.05 are marked in bold.}
\center
\begin{tabular}{@{}l l l l@{}}
\toprule
\multirow{2}{*}{Data} & \multirow{2}{*}{Method, $v_{R^2}$} & \multicolumn{2}{c}{Test $R^2$}\\
  &   & 50\%,\ \ (25\%,\ 75\%) & p-value\\
\midrule
\multirow{7}{*}{CPU Run Time}
& KGD-D, 0.02 & $0.80,\ (0.65, 0.89)$ & $\bm{4.7e\text{-}7}$\\
& KGD-D, 0.05 & $0.79,\ (0.62, 0.89)$ & $\bm{2.1e\text{-}5}$\\
& KGD-D, 0.1  & $0.77,\ (0.58, 0.87)$ & $\bm{0.0055 }$\\
& KGD-D, 0.2  & $0.69,\ (0.53, 0.84)$ & $0.17   $\\
& KGD-D, 0.5  & $0.64,\ (0.41, 0.80)$ & $0.88   $\\
\cline{2-4}
& KRR-CV      & $0.68,\ (0.42, 0.86)$ & $-$\\
& KRR-MML     & $0.49,\ (0.13, 0.67)$ & $-$\\
\hline
\multirow{7}{*}{Power Consumption}
& KGD-D, 0.02 & $0.61,\ (0.47, 0.72)$ & $0.94   $\\
& KGD-D, 0.05 & $0.65,\ (0.52, 0.75)$ & $\bm{5.9e\text{-}5}$\\
& KGD-D, 0.1  & $0.67,\ (0.55, 0.76)$ & $\bm{5.5e\text{-}10}$\\
& KGD-D, 0.2  & $0.66,\ (0.55, 0.75)$ & $\bm{9e\text{-}7  }$\\
& KGD-D, 0.5  & $0.61,\ (0.49, 0.69)$ & $1      $\\
\cline{2-4}
& KRR-CV      & $0.62,\ (0.46, 0.73)$ & $-$\\
& KRR-MML     & $0.53,\ (0.40, 0.63)$ & $-$\\
\hline
\multirow{7}{*}{Superconductors}
& KGD-D, 0.02 & $0.64,\ (0.52, 0.74)$ & $1      $\\
& KGD-D, 0.05 & $0.67,\ (0.56, 0.76)$ & $\bm{0.02   }$\\
& KGD-D, 0.1  & $0.68,\ (0.57, 0.76)$ & $\bm{7.4e\text{-}6}$\\
& KGD-D, 0.2  & $0.67,\ (0.56, 0.74)$ & $\bm{0.00042}$\\
& KGD-D, 0.5  & $0.62,\ (0.52, 0.69)$ & $1      $\\
\cline{2-4}
& KRR-CV      & $0.65,\ (0.54, 0.74)$ & $-$\\
& KRR-MML     & $0.57,\ (0.46, 0.64)$ & $-$\\
\hline
\multirow{7}{*}{U.K.\ Temperature}
& KGD-D, 0.02 & $0.46,\ (0.31, 0.58)$ & $1      $\\
& KGD-D, 0.05 & $0.52,\ (0.40, 0.63)$ & $0.97   $\\
& KGD-D, 0.1  & $0.54,\ (0.44, 0.64)$ & $\bm{0.011  }$\\
& KGD-D, 0.2  & $0.55,\ (0.44, 0.63)$ & $\bm{0.00073}$\\
& KGD-D, 0.5  & $0.49,\ (0.39, 0.58)$ & $1      $\\
\cline{2-4}
& KRR-CV      & $0.52,\ (0.42, 0.61)$ & $-$\\
& KRR-MML     & $0.46,\ (0.37, 0.53)$ & $-$\\
\bottomrule
\end{tabular}
\label{tab:dec_100}
\end{table}

In Table \ref{tab:dec_100}, we compare the performance of KGD-D with $\sigma_m=0$, for five different values of $v_{R^2}$, to KRR with constant bandwidth, where the bandwidth and regularization strengths are selected by 10-fold cross-validation, KRR-CV, and marginal likelihood maximization, KRR-MML. 
For $v_{R^2}=0.1$, KGD-D performs significantly better than KRR-CV and KRR-MML, in terms of test $R^2$, on all four data sets.
Each experiment took approximately 0.1 seconds for KGD-D and KRR-MML and 1 second for KRR-CV, to run on an Intel i9-13980HX processor with access to 32 GB RAM.

\section{Conclusions}
We showed, theoretically and on real and synthetic data, that decreasing the bandwidth of a translational-invariant kernel during training, leads to a double descent behavior in the model complexity, expressed as the minimum allowed bandwidth.

Decreasing the bandwidth all the way to zero leads to benign overfitting, circumvents the need for model selection, and outperforms kernel regression with a constant bandwidth, selected by cross-validation or marginal likelihood maximization.

Preliminary results show that using preconditioned gradient descent to make the NTK evolve as if its bandwidth were decreasing during training, we can make neural networks converge in fewer iterations, to a solution that overfits more benignly, compared to standard gradient descent. Here, however, further research is needed.

\newpage
\clearpage
\newpage
\clearpage
\newpage
\bibliography{refs}
\bibliographystyle{apalike}

\newpage
\appendix
\onecolumn
\section{Complexity as Function of Bandwidth}
\label{sec:bw_compl}
A common way to characterize the complexity of a model is by its effective number of parameters, also referred to as the effective degrees of freedom, df. In Proposition \ref{thm:sigma_compl}, we state the effective number of parameters, and the model estimates, for zero and infinite bandwidths for KRR with a translational-invariant kernel. In the absence of regularization, the effective number of parameters is 1 for infinite bandwidth and $n$ for zero bandwidth. Since, $\df=[0,n]$, Equation \ref{eq:sigma_df} suggests that $\sigma=\infty$ corresponds to the simplest possible model, with only one parameter, while $\sigma=0$ corresponds to the most complex possible model. This is supported by Equation \ref{eq:sigma_fun}, according to which, in the absence of regularization, the inferred model for $\sigma=\infty$ is simply the mean of the observations, while, for $\sigma=0$, the inferred function predicts zero (the prior) everywhere, except at the training observations, where the prediction is the observation, arguably most complex function possible.

\begin{prop}~\\
\label{thm:sigma_compl}
Let $k(\xv,\xpv,\sigma)=k\left(\frac{\|\xv-\xpv\|_2}{\sigma}\right)$ be a translational-invariant kernel such that $k(0)=1$, $k(\infty)=0$ and let $\fh(\xsv,\X,\lambda,\sigma)$ denote the KRR estimate for $\xsv$.
Then, 

\begin{equation}
\label{eq:sigma_df}
\begin{aligned}
&\df\left(\fh(\xsv,\X,\lambda,\sigma=\infty)\right)=\frac1{1+\lambda/n}\\
&\df\left(\fh(\xsv,\X,\lambda,\sigma=0)\right)=\frac{n}{1+\lambda},
\end{aligned}
\end{equation}
and
\begin{equation}
\label{eq:sigma_fun}
\begin{aligned}
&\fh(\xsv,\X,\lambda,\sigma=\infty)=\frac1{1+\lambda/n}\cdot \frac1n\sum_{i=1}^ny_i\\
&\fh(\xsv,\X,\lambda,\sigma=0)=
\begin{cases}
0&\text{ if } \xsv \notin \X\\
\frac1{1+\lambda}\cdot y_i&\text{ if } \xsv = \xvi\in \X.
\end{cases}
\end{aligned}
\end{equation}
\end{prop}

Another perspective on relating bandwidth and model complexity is to use the derivatives of the inferred function as a complexity measure. Intuitively, restricting the derivatives of a function restricts its complexity, and this is also what is used in many regularization techniques. The simplest example is probably penalized linear regression: For $\fh(\xsv)=\xsv{^\top}\bm{\hat{\beta}}$, $\partial_{\xsv}\fh(\xsv)=\bm{\hat{\beta}}$, and thus parameter shrinking techniques, such as ridge regression and lasso, constrain the function complexity by restricting its derivatives. Other regression techniques where the function complexity is restricted through the derivatives are Jacobian regularization \citep{jakubovitz2018improving} for neural networks, and smoothing splines (in the second case through the second derivative).
According to Proposition \ref{thm:der_bound_lbda}, for a fixed training time, the gradient of the inferred function is bounded by $1/\sigma$. Just as Proposition \ref{thm:sigma_compl}, Proposition \ref{thm:der_bound_lbda} suggests that a model with a larger bandwidth results in a less complex inferred function, and additionally suggests that the relation between complexity and bandwidth is the multiplicative inverse.

\begin{prop}~\\
\label{thm:der_bound_lbda}
Let $k(\xv,\xpv,\sigma)=k\left(\|\xv-\xpv\|_2/\sigma\right)$ be a translational-invariant kernel with bounded derivative, $\max_{u\geq 0}|k'(u)|= k'_{\max}<\infty$. Then the derivative of the KRR estimate, $\fh(\xsv,\X,\lambda,\sigma)$, is bounded according to
\begin{equation*}
\label{eq:der_bound}
\left\|\frac{\partial \fh(\xsv,\X,\lambda,\sigma)}{\partial \xsv}\right\|_2 \leq \frac1\sigma\cdot \frac1{\max(\lambda,\ \smin(\K))}\cdot\|\yv\|_2\cdot k'_{\max}\cdot\sqrt{n}.
\end{equation*}
where $\smin(\K)$ denotes the smallest singular value of $\K$.
\end{prop}

\section{Experimental Details}
\label{sec:exp_dets}
\subsection{Details for Figure \ref{fig:syn_dd_100}}
The true function is given by $y=\sin(2\pi x)$. We sampled $n=20$ training observations according to $x_i\sim \mathcal{U}(-1,1)$, $y_i\sim\mathcal{N}(\sin(2\pi x), 0.2^2)$, where $\mathcal{U}(a,b)$ denotes the uniform distribution on $[a,b]$, and $\mathcal{N}(\mu,\varsigma^2)$ denotes the normal distribution with mean $\mu$ and variance $\varsigma^2$. In addition, we used 100 uniformly spaced out-of-sample data points on the interval [-1,1]. 
We used the Gaussian kernel, $k(\xv,\xpv)=\exp\left(-\frac{\|\xv-\xpv\|^2_2}{2\sigma^2}\right)$ for both KGD-D and KRR. 
For KGD-D, we used $\Delta t=0.01$, and $\sigma_0=\max_{i,j}\|\xvi-\xvj\|_2$, where $\xvi$ and $\xvj$ are rows in $\X$, and $v_{R^2}=0.1$. We used a fixed training time/regularization of $\lambda=1/t=0.001$.
For the bottom panel, we used 100 different values of $\sigma_m$ and 100 different realizations to obtain the prediction intervals.

\subsection{Details for Figure \ref{fig:syn_nn}}
The first four neural networks have 1, 1, 2, and 4 hidden layers, with 1, 3, 200, and 200 nodes respectively. The fifth network, trained with preconditioned gradient descent, has one hidden layer with 50 nodes. We used the tanh activation function for all hidden layers. For the first four models, we used a learning rate of 0.001 and a momentum of 0.95. For the preconditioned gradient descent model, we used the same learning rate of 0.001, no momentum (for stability reasons), and $v_{R^2}=0.1$.

\subsection{Details for Section \ref{sec:exps}}
In all experiments, the data were standardized to zero mean and unit variance. 
For KGD-D, we used $\Delta t=0.01$, and $\sigma_0=\max_{i,j}\|\xvi-\xvj\|_2$, where $\xvi$ and $\xvj$ are rows in $\X$, and, except when otherwise is stated in Tables \ref{tab:dec_100}, \ref{tab:dec_0.5}, and \ref{tab:dec_2.5}, $v_{R^2}=0.1$. 

For Figure \ref{fig:dd_100}, we used 100 different values of $\sigma_m$ and 100 random, non-overlapping splits, each containing 100 observations (for the smaller CPU data set, 93 of the splits were of size 82 and 7 of size 81) to obtain the prediction intervals.
Each experiment took approximately 2 seconds on an Intel i9-13980HX processor with access to 32 GB RAM.

For Table \ref{tab:dec_100}, we used 200 random, non-overlapping splits, each containing 100 observations (for the smaller CPU data set, we instead used 100 splits, 93 of them of size 82 and 7 of size 81), which in turn were split 80\%/20\% into training and test data.
For KRR-CV, we used $30\times30$ logarithmically spaced values for $\lambda\in [10^{-6},1]$ and $\sigma\in[0.01,10]$. For KRR-MML, to mitigate the problem of convergence to a poor local optimum, $5\times 5$ different logarithmically spaced optimization seeds were used for $\lambda$ and $\sigma$. 

The CPU Run Time data are available at\\\url{http://www.cs.toronto.edu/~delve/data/comp-activ/desc.html}. We excluded the features \texttt{usr}, \texttt{sys}, \texttt{wio}, and \texttt{idle}.\\
The Power Consumption data are available at\\\url{https://archive.ics.uci.edu/dataset/849/power+consumption+of+tetouan+city}\\under the CC-BY 4.0 license. We excluded the \texttt{DateTime} feature, using the remaining features to predict \texttt{Zone 3 Power Consumption}.\\
The Superconductor data are available at\\\url{https://archive.ics.uci.edu/dataset/464/superconductivty+data}\\under the CC-BY 4.0 license.\\
The U.K.\ temperature data are available at \url{https://www.maths.ed.ac.uk/~swood34}\\under licenses from the OS Open Data and UK Met Office. We only used data for the year 2000. We used the features \texttt{yday}, \texttt{x}, \texttt{y}, and \texttt{h} to predict \texttt{Tmean1}.

The total run time, for all experiments, was approximately 30 hours. To decrease the wall-clock time, we used a cluster with AMD EPYC Zen 2, 2.25 GHz cores, using one core per experiment.

\section{KGD-D as Gradient Boosting}
\label{sec:fsam}
Gradient boosting \citep{friedman2001greedy} is a form of boosting where $m$ consecutive models are used, and where model $m$ is used to fit the residuals of model $m-1$, i.e.\ 
\begin{equation*}
\begin{aligned}
\fhv_m'(\X)&=\argmin_{\fv'_m\in\mathcal{F}_m}L\left(\fv'_m(\X),\yv-\fhv_{m-1}(\X)\right)\\
\fhv_m(\X)&=\fhv_{m-1}(\X)+\fhv_m'(\X),
\end{aligned}
\end{equation*}
where $\mathcal{F}_m$ is some class of functions and $L$ is a loss function quantifying the discrepancy between $\fv'_m$ and $\yv-\fhv_{m-1}$. 

In general, the same class, $\mathcal{F}_m=\mathcal{F}$, is used in all stages and the complexity of $\mathcal{F}$ has to be carefully selected to obtain good performance. However, if the complexity of $\mathcal{F}_m$ is allowed to increase with $m$, then simpler relations in the data will be captured first, by the simpler models, while more complex relations will be captured in later stages. Thus (ideally) each part of the data is modeled by a model of exactly the required complexity. Furthermore, if a simple model is enough to model the data, there will be no residuals left to fit for the more complex models and the total model will not be more complex than needed.

Gradient descent is an iterative algorithm, where the update in each iteration is based on the output of the previous iteration. Thus, any iteration can be thought of as starting the training of a new model from the beginning, using the output of the previous iteration as a warm start. If this interpretation is made every time the bandwidth is changed, then KGD-D becomes exactly gradient boosting, where the different models are defined by their bandwidths. Thus, if the bandwidth is updated during training in such a way that the model complexity increases during training, we obtain exactly gradient boosting with increasing complexity.

\begin{figure}
\center
\includegraphics[width=\textwidth]{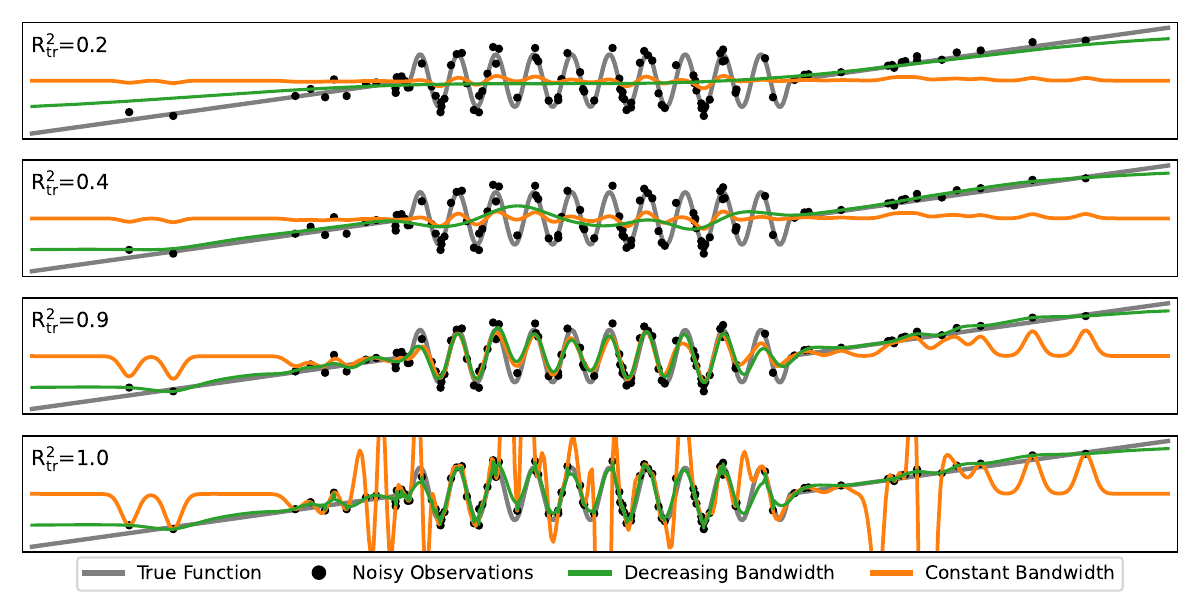}
\caption{Inferred models at different training times, increasing with the row number. Training time/regularization is selected to obtain the same $\Rttr\in[0,1]$ for both models. KGD-D first captures the linear parts of the data, and performs well both on the linear and sinusoidal parts, in combination with perfectly interpolating the data at convergence. In contrast, the constant bandwidth (selected by 10-fold cross-validation) is too small to perform well on the linear parts and too large to generalize well when the training error goes to zero.}
\label{fig:inc_compl_demo}
\end{figure}

In Figure \ref{fig:inc_compl_demo}, we demonstrate how KGD-D uses functions of increasing complexity during training, in contrast to KRR with constant bandwidth. For KGD-D, the model complexity increases with training time, from almost linear to perfectly interpolating the training data. KRR, in contrast, uses a bandwidth that is too small to perform well on the linear parts and too large to generalize well when the training error goes to zero.

\section{KGD-D Inspired Preconditioned Gradient Descent for Neural Networks}
\label{sec:ntk}
Updating the parameters of a neural network (or actually any iteratively trained regression model) with parameters $\thetahv\in\R^p$ and loss function $L(\fhv,\yv)$, trained with gradient descent, amounts to
\begin{equation*}
\thetahv(t+\Delta t)=\thetahv(t)-\Delta t\cdot\frac{\partial L(\fhv(\thetahv(t)),\yv)}{\partial \thetahv(t)} =\thetahv(t)-\Delta t\cdot\left(\frac{\partial \fhv(\thetahv(t))}{\partial \thetahv(t)}\right)^\top\cdot\frac{\partial L(\fhv(\thetahv(t)),\yv)}{\partial \fhv(\thetahv(t))},
\end{equation*}
where we have used the chain rule, and where $\Delta t$ denotes the learning rate. Rearranging and letting $\Delta t\to 0$, we obtain the gradient flow update for $\thetahv$, 
\begin{equation*}
\frac{\partial \thetahv(t)}{\partial t}=-\left(\frac{\partial \fhv(\thetahv(t))}{\partial \thetahv(t)}\right)^\top\cdot\frac{\partial L(\fhv(\thetahv(t)),\yv)}{\partial \fhv(\thetahv(t))},
\end{equation*}
and with another application of the chain rule, the gradient flow update for $[\fhv^\top, \fhsv{^\top}]^\top$ is obtained as

\begin{equation*}
\begin{aligned}
\frac{\partial}{\partial t}\left(\begin{bmatrix}\fhv(\thetahv(t))\\\fhsv(\thetahv(t))\end{bmatrix}\right)
&=\begin{bmatrix}\partial_{\thetahv(t)}\fhv(\thetahv(t))\\\partial_{\thetahv(t)}\fhsv(\thetahv(t))\end{bmatrix}\cdot\frac{\partial \thetahv(t)}{\partial t}\\
&=-\begin{bmatrix}\partial_{\thetahv(t)}\fhv(\thetahv(t))\\\partial_{\thetahv(t)}\fhsv(\thetahv(t))\end{bmatrix}\cdot\left(\partial_{\thetahv(t)} \fhv(\thetahv(t))\right)^\top\cdot\frac{\partial L(\fhv(\thetahv(t)),\yv)}{\partial \fhv(\thetahv(t))}\\
&=:-\Kntk(\thetahv(t))\cdot\frac{\partial L(\fhv(\thetahv(t)),\yv)}{\partial \fhv(\thetahv(t))},
\end{aligned}
\end{equation*}
where
\begin{equation*}
\Kntk(t):= \begin{bmatrix}\partial_{\thetahv}\fhv(t)\\\partial_{\thetahv}\fhsv(t)\end{bmatrix}\cdot\left(\partial_{\thetahv} \fhv(t)\right)^\top=\bm{\Phi}(t)\bm{\Phi_t}(t)^\top,
\end{equation*}
for $\bm{\Phi}:=\begin{bmatrix}\partial_{\thetahv}\fhv\\\partial_{\thetahv}\fhsv\end{bmatrix}$, $\bm{\Phi_t}:=\partial_{\thetahv} \fhv$.

In preconditioned gradient descent, the gradient is multiplied by a preconditioning matrix, $\bm{P}$, changing the expressions above into

\begin{equation*}
\begin{aligned}
&\thetahv(t+\Delta t)=\thetahv(t)-\Delta t\cdot\bm{P}(t)\cdot\frac{\partial L(\fhv(t),\yv)}{\partial \thetahv}\\
&\frac{\partial \thetahv}{\partial t}=-\bm{P}(t)\cdot\left(\frac{\partial \fhv(t)}{\partial \thetahv}\right)^\top\cdot\frac{\partial L(\fhv(t),\yv)}{\partial \fhv}\\
&\frac{\partial}{\partial t}\left(\begin{bmatrix}\fhv(t)\\\fhsv(t)\end{bmatrix}\right)
=-\bm{\Phi}(t)\cdot\bm{P}(t)\cdot\bm{\Phi_t}(t)^\top\cdot\frac{\partial L(\fhv(t),\yv)}{\partial \fhv}
=:-\KntkP(t)\cdot\frac{\partial L(\fhv(t),\yv)}{\partial \fhv},
\end{aligned}
\end{equation*}
where
\begin{equation*}
\KntkP(t):= \bm{\Phi}(t)\cdot\bm{P}(t)\cdot\bm{\Phi_t}(t)^\top.
\end{equation*}

Assume that $p>n+\ns$ and let $\bm{\Phi}^+:=\bm{\Phi}^\top(\bm{\Phi}\bm{\Phi}^\top)^{-1}$ and $\bm{\Phi_t}^+:=\bm{\Phi_t}^\top(\bm{\Phi_t}\bm{\Phi_t}^\top)^{-1}$ denote the pseudo-inverses of $\bm{\Phi}$ and $\Ph_t$. Then $\bm{\Phi}\bm{\Phi}^+$ and $\Pht\Pht^+$ are both identity matrices. Now, if

\begin{equation*}
\bm{P}(t)=\Ph(t)^+\bm{\tilde{K}}(t)\Pht(t)^{+\top},
\end{equation*}
for some kernel matrix $\bm{\tilde{K}}\in\R^{(n+\ns)\times n}$, then
\begin{equation*}
\begin{aligned}
\frac{\partial}{\partial t}\left(\begin{bmatrix}\fhv(t)\\\fhsv(t)\end{bmatrix}\right)
&=-\bm{\Phi}(t)\Ph(t)^+\cdot\bm{\tilde{K}}(t)\cdot\left(\bm{\Phi_t}(t)\Pht(t)^+\right)^\top\cdot\frac{\partial L(\fhv(t),\yv)}{\partial \fhv}\\
&=-\bm{\tilde{K}}(t)\cdot\frac{\partial L(\fhv(t),\yv)}{\partial \fhv},
\end{aligned}
\end{equation*}
which means that we can replace the NTK with any kernel, albeit at the cost of computing the pseudo-inverse of the NTK feature expansion, which has computational complexity $p\cdot(\ns+n)^2$.
In Section \ref{sec:nn} we simply used 
\begin{equation*}
\bm{\tilde{K}}(t)=\alpha(t)\cdot\Kntk(t)+(1-\alpha(t))\cdot\begin{bmatrix}\I\\\nv\end{bmatrix},
\end{equation*}
decreasing $\alpha$ from one to zero during training, thus mimicking the effect of a bandwidth going to zero.

\section{Additional Experiments}
\label{sec:more_exps}
In this section, we repeat the experiments of Figure \ref{fig:syn_dd_100} and Section \ref{sec:exps} for the four kernels in Table \ref{tab:kerns}, and extend Table \ref{tab:dec_100} with the results of KGF with cross-validation, KGF-CV, and a multilayer perceptron, MLP. Since the connection to Gaussian process regression is lost for KGF, we cannot use marginal likelihood maximization in this case. KGF-CV is implemented identically to KRR-CV, but with Equation \ref{eq:krr_s} replaced by Equation \ref{eq:kgf_s} for $t=1/\lambda$.
For the MLP, we used two hidden layers, each with 100 nodes and the tanh activation function, and a learning rate of 0.01, setting aside 10\% of the training data to validate when to stop the training. 

\begin{figure}
\center
\includegraphics[width=\textwidth]{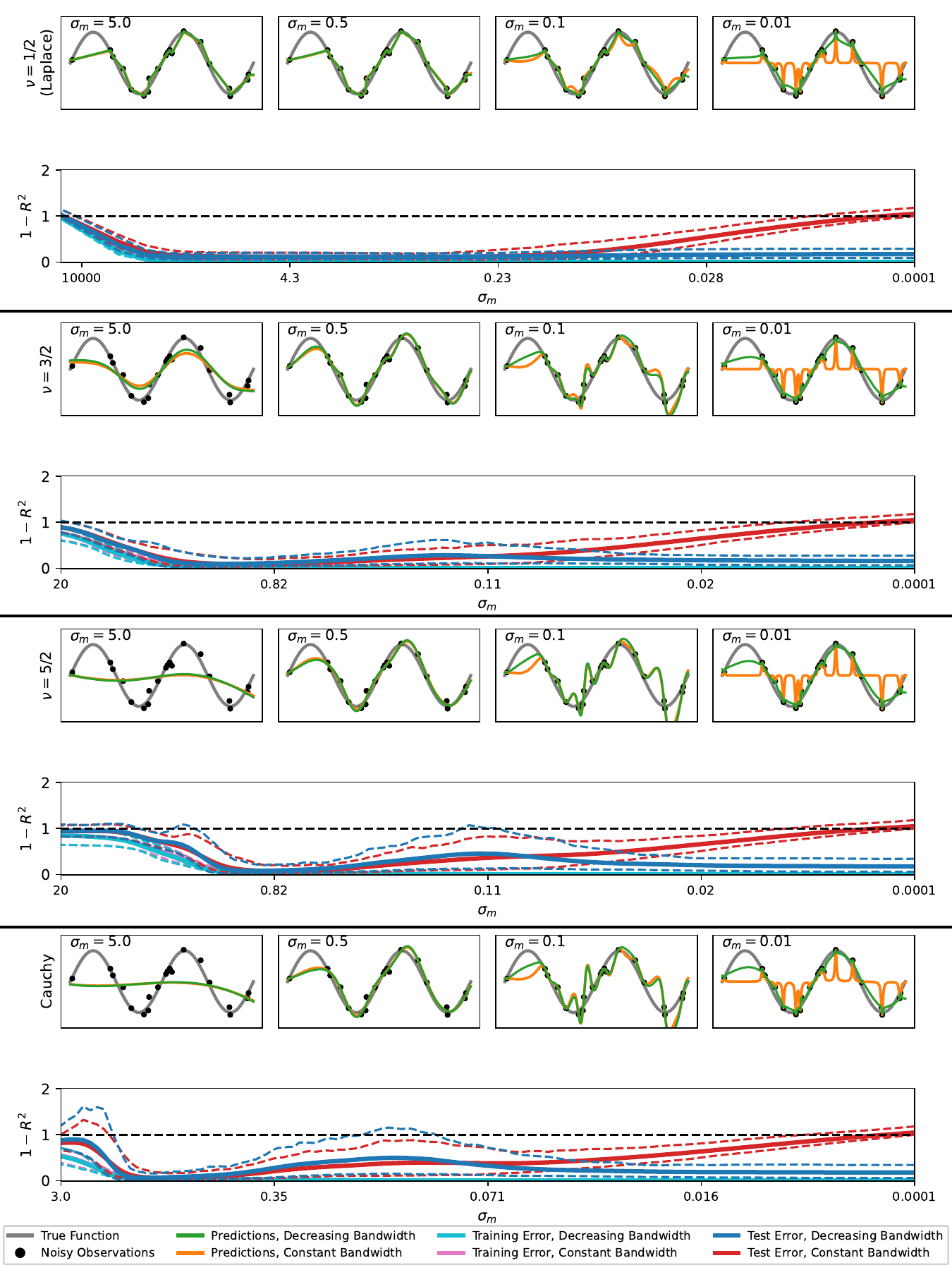}
\caption{Analogue of Figure \ref{fig:syn_dd_100}, for the four kernels in Table \ref{tab:kerns}. Except for $\nu=1/2$, for which the kernel and inferred functions tend to linearly interpolate the data, the results agree with those of Figure \ref{fig:dd_100}.}
\label{fig:syn_dd_more}
\end{figure}

\begin{figure}
\center
\includegraphics[width=\textwidth]{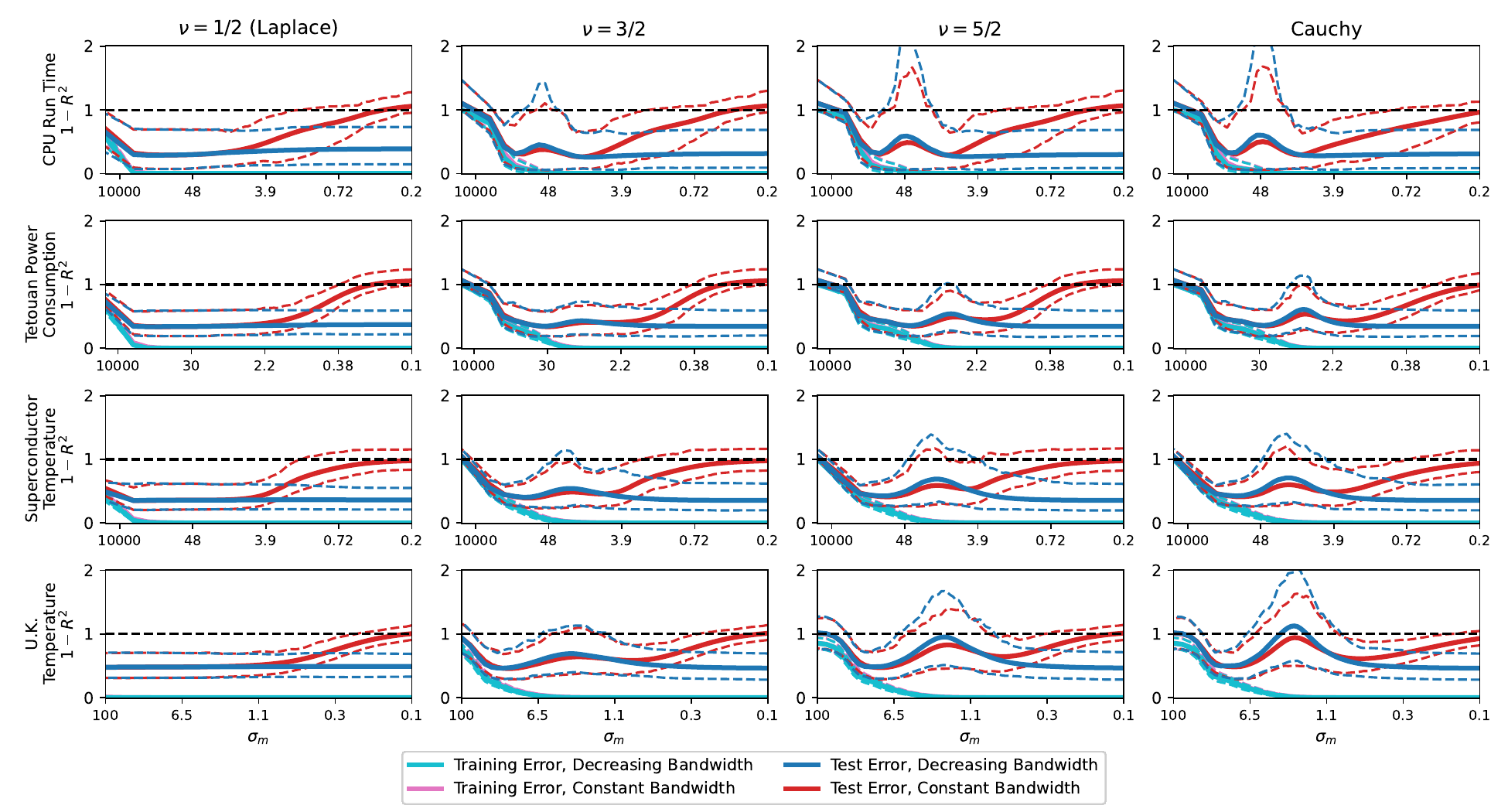}
\caption{Training and test errors (as $1-R^2$) as functions of model complexity (as in terms of $\sigma_m$), for kernel regression with constant and decreasing bandwidths. The plots show the means, together with the 90\% prediction intervals. Except for $\nu=1/2$, for which the kernel and inferred functions tend to linearly interpolate the data, the results agree with those of Figure \ref{fig:dd_100}.}
\label{fig:dd_more}
\end{figure}

The analogues of Figures \ref{fig:syn_dd_100} and \ref{fig:dd_100} are presented in Figures \ref{fig:syn_dd_more} and \ref{fig:dd_more}, and the analogues of Table \ref{tab:dec_100} in Tables \ref{tab:dec_100_ext}, \ref{tab:dec_0.5} and \ref{tab:dec_2.5}. The results are confirmed with one exception: As seen in Figures \ref{fig:syn_dd_more} and \ref{fig:dd_more}, the Laplace kernel does not exhibit a double descent behavior, instead the constant and decreasing bandwidth schemes exhibit the same generalization error for a large fraction of the $\sigma_m$s. This is also confirmed in Table \ref{tab:dec_0.5}, where KRR-CV performs similarly to KGD-D. This can probably be attributed to the fact that for the Laplace kernel, all derivatives of the kernel, and thus of the inferred function, are discontinuous. As can be seen in Figure \ref{fig:syn_dd_more}, this kernel tends to perform linear interpolation between the observations, which reduces the impact of the constant bandwidth. However, KGD-D with the Gaussian kernel always outperforms KRR-CV with the Laplace kernel.

KGF-CV and KRR-CV perform virtually identically. KGD-D tends to perform slightly better than the MLP.
Each experiment took approximately 2 seconds for the MLP and 20 seconds for KGF-CV to run on an Intel i9-13980HX processor with access to 32 GB RAM.

\begin{table}
\caption{Additional kernels used.}
\center
\begin{tabular}{@{}l l@{}}
\toprule
Name & Equation \\
\midrule
Matérn, $\nu=\frac12$ (Laplace) & $\exp\left(-\frac{\|\xv-\xpv\|_2}\sigma\right)$\\
Matérn, $\nu=\frac32$  & $\left(1+\frac{\sqrt{3}\cdot\|\xv-\xpv\|_2}\sigma\right)\cdot\exp\left(-\frac{\sqrt{3}\cdot\|\xv-\xpv\|_2}\sigma\right)$\\
Matérn, $\nu=\frac52$  & $\left(1+\frac{\sqrt{5}\cdot\|\xv-\xpv\|_2}\sigma+\frac{5\cdot\|\xv-\xpv\|^2_2}{3\cdot\sigma^2}\right)\cdot\exp\left(-\frac{\sqrt{5}\cdot\|\xv-\xpv\|_2}\sigma\right)$\\
Cauchy & $\left(1+\frac{\|\xv-\xpv\|_2^2}{\sigma^2}\right)^{-1}$\\
\bottomrule
\end{tabular}
\label{tab:kerns}
\end{table}

\begin{table}
\center
\caption{Median and first and third quartile, over the 200 (or 100 for the CPU data) splits, of $R^2$ on the test data. For KGF-CV, we used the Gaussian kernel.}
\begin{tabular}{@{}l l l@{}}
\toprule
Data & Method & Test $R^2$, 50\%,\ \ (25\%,\ 75\%) \\
\midrule
\multirow{2}{*}{CPU Run Time}
& KGF-CV      & $0.67,\ (0.36, 0.84)$ \\
& MLP         & $0.67,\ (0.49, 0.83)$ \\
\hline
\multirow{2}{*}{Power Consumption}
& KGF-CV      & $0.61,\ (0.44, 0.72)$ \\
& MLP         & $0.31,\ (0.15, 0.44)$ \\
\hline
\multirow{2}{*}{Superconductors}
& KGF-CV      & $0.64,\ (0.52, 0.74)$ \\
& MLP         & $0.54,\ (0.34, 0.66)$ \\
\hline
\multirow{2}{*}{U.K.\ Temperature}
& KGF-CV      & $0.51,\ (0.42, 0.62)$ \\
& MLP         & $0.29,\ (0.047, 0.49)$ \\
\bottomrule
\end{tabular}
\label{tab:dec_100_ext}
\end{table}


\begin{table}
\caption{Median and first and third quartile, over the 200 (or 100 for the CPU data) splits, of $R^2$ on the test data. The p-value is that of a Wilcoxon signed-rank test, testing whether the method performs better than KRR-CV and KRR-MML, where p-values less than 0.05 are marked in bold.}
\center
\begin{tabular}{@{}l l l l l l@{}}
\toprule
\multirow{2}{*}{Data} & \multirow{2}{*}{Method, $v_{R^2}$} & \multicolumn{2}{c}{Test $R^2$, $\nu=1/2$ (Laplace)} & \multicolumn{2}{c}{Test $R^2$, $\nu=3/2$}\\
  &   & 50\%,\ \ (25\%,\ 75\%) & p-value & 50\%,\ \ (25\%,\ 75\%) & p-value\\
\midrule
\multirow{8}{*}{CPU Run Time}
& KGD-D, 0.02 & $0.70,\ (0.50, 0.81)$ & $\bm{0.04   }$ & $0.79,\ (0.62, 0.88)$ & $\bm{0.00078}$\\
& KGD-D, 0.05 & $0.68,\ (0.48, 0.80)$ & $0.89   $ & $0.77,\ (0.60, 0.87)$ & $0.13   $\\
& KGD-D, 0.1  & $0.66,\ (0.47, 0.79)$ & $1      $ & $0.73,\ (0.54, 0.85)$ & $0.93   $\\
& KGD-D, 0.2  & $0.63,\ (0.43, 0.77)$ & $1      $ & $0.69,\ (0.47, 0.82)$ & $1      $\\
& KGD-D, 0.5  & $0.54,\ (0.34, 0.74)$ & $1      $ & $0.61,\ (0.40, 0.78)$ & $1      $\\
\cline{2-6}
& KRR-CV      & $0.71,\ (0.47, 0.82)$ & $-$ & $0.77,\ (0.56, 0.86)$ & $-$\\
& KRR-MML     & $0.46,\ (0.19, 0.66)$ & $-$ & $0.50,\ (0.17, 0.68)$ & $-$\\
& KGF-CV      & $0.71,\ (0.47, 0.82)$ & $-$ & $0.77,\ (0.55, 0.85)$ & $-$\\
\hline
\multirow{8}{*}{Power Consumption}
& KGD-D, 0.02 & $0.65,\ (0.55, 0.74)$ & $0.87   $ & $0.66,\ (0.52, 0.75)$ & $0.84   $\\
& KGD-D, 0.05 & $0.64,\ (0.53, 0.73)$ & $1      $ & $0.67,\ (0.55, 0.75)$ & $\bm{0.014  }$\\
& KGD-D, 0.1  & $0.64,\ (0.52, 0.72)$ & $1      $ & $0.67,\ (0.55, 0.75)$ & $\bm{0.0088 }$\\
& KGD-D, 0.2  & $0.61,\ (0.51, 0.70)$ & $1      $ & $0.64,\ (0.54, 0.74)$ & $0.87   $\\
& KGD-D, 0.5  & $0.55,\ (0.45, 0.63)$ & $1      $ & $0.60,\ (0.48, 0.67)$ & $1      $\\
\cline{2-6}
& KRR-CV      & $0.65,\ (0.54, 0.74)$ & $-$ & $0.65,\ (0.52, 0.74)$ & $-$\\
& KRR-MML     & $0.50,\ (0.41, 0.59)$ & $-$ & $0.52,\ (0.42, 0.62)$ & $-$\\
& KGF-CV      & $0.65,\ (0.53, 0.74)$ & $-$ & $0.65,\ (0.51, 0.73)$ & $-$\\
\hline
\multirow{8}{*}{Superconductors}
& KGD-D, 0.02 & $0.68,\ (0.58, 0.76)$ & $\bm{7e\text{-}10  }$ & $0.67,\ (0.55, 0.76)$ & $0.92   $\\
& KGD-D, 0.05 & $0.67,\ (0.57, 0.76)$ & $\bm{3e\text{-}8  }$ & $0.68,\ (0.57, 0.77)$ & $\bm{0.02   }$\\
& KGD-D, 0.1  & $0.67,\ (0.56, 0.75)$ & $\bm{0.0016 }$ & $0.68,\ (0.57, 0.76)$ & $\bm{0.00066}$\\
& KGD-D, 0.2  & $0.65,\ (0.55, 0.74)$ & $0.97   $ & $0.67,\ (0.56, 0.75)$ & $0.37   $\\
& KGD-D, 0.5  & $0.59,\ (0.50, 0.67)$ & $1      $ & $0.61,\ (0.52, 0.70)$ & $1      $\\
\cline{2-6}
& KRR-CV      & $0.66,\ (0.55, 0.75)$ & $-$ & $0.67,\ (0.55, 0.75)$ & $-$\\
& KRR-MML     & $0.56,\ (0.45, 0.64)$ & $-$ & $0.57,\ (0.47, 0.65)$ & $-$\\
& KGF-CV      & $0.66,\ (0.56, 0.75)$ & $-$ & $0.66,\ (0.55, 0.75)$ & $-$\\
\hline
\multirow{8}{*}{U.K.\ Temperature}
& KGD-D, 0.02 & $0.53,\ (0.42, 0.64)$ & $\bm{5.9e\text{-}5}$ & $0.52,\ (0.41, 0.63)$ & $1      $\\
& KGD-D, 0.05 & $0.53,\ (0.42, 0.63)$ & $\bm{0.015  }$ & $0.53,\ (0.43, 0.64)$ & $0.34   $\\
& KGD-D, 0.1  & $0.51,\ (0.40, 0.62)$ & $0.94   $ & $0.55,\ (0.44, 0.64)$ & $\bm{0.012  }$\\
& KGD-D, 0.2  & $0.50,\ (0.39, 0.59)$ & $1      $ & $0.54,\ (0.43, 0.63)$ & $0.62   $\\
& KGD-D, 0.5  & $0.42,\ (0.32, 0.50)$ & $1      $ & $0.48,\ (0.36, 0.55)$ & $1      $\\
\cline{2-6}
& KRR-CV      & $0.52,\ (0.42, 0.62)$ & $-$ & $0.54,\ (0.43, 0.64)$ & $-$\\
& KRR-MML     & $0.37,\ (0.29, 0.43)$ & $-$ & $0.42,\ (0.34, 0.49)$ & $-$\\
& KGF-CV      & $0.53,\ (0.43, 0.63)$ & $-$ & $0.53,\ (0.43, 0.64)$ & $-$\\
\bottomrule
\end{tabular}
\label{tab:dec_0.5}
\end{table}

\begin{table}
\caption{Median and first and third quartile, over the 200 (or 100 for the CPU data) splits, of $R^2$ on the test data. The p-value is that of a Wilcoxon signed-rank test, testing whether the method performs better than KRR-CV and KRR-MML, where p-values less than 0.05 are marked in bold.}
\center
\begin{tabular}{@{}l l l l l l@{}}
\toprule
\multirow{2}{*}{Data} & \multirow{2}{*}{Method, $v_{R^2}$} & \multicolumn{2}{c}{Test $R^2$, $\nu=5/2$} & \multicolumn{2}{c}{Test $R^2$, Cauchy}\\
  &   & 50\%,\ \ (25\%,\ 75\%) & p-value & 50\%,\ \ (25\%,\ 75\%) & p-value\\
\midrule
\multirow{8}{*}{CPU Run Time}
& KGD-D, 0.02 & $0.81,\ (0.65, 0.88)$ & $\bm{0.00044}$ & $0.80,\ (0.63, 0.88)$ & $\bm{0.00028}$\\
& KGD-D, 0.05 & $0.80,\ (0.60, 0.88)$ & $\bm{0.015  }$ & $0.78,\ (0.59, 0.88)$ & $\bm{0.028  }$\\
& KGD-D, 0.1  & $0.76,\ (0.56, 0.86)$ & $0.34   $ & $0.74,\ (0.54, 0.85)$ & $0.51   $\\
& KGD-D, 0.2  & $0.70,\ (0.50, 0.82)$ & $0.91   $ & $0.69,\ (0.48, 0.82)$ & $0.99   $\\
& KGD-D, 0.5  & $0.63,\ (0.41, 0.78)$ & $1      $ & $0.61,\ (0.41, 0.78)$ & $1      $\\
\cline{2-6}
& KRR-CV      & $0.74,\ (0.51, 0.85)$ & $-$ & $0.74,\ (0.53, 0.84)$ & $-$\\
& KRR-MML     & $0.50,\ (0.16, 0.67)$ & $-$ & $0.50,\ (0.21, 0.68)$ & $-$\\
& KGF-CV      & $0.74,\ (0.51, 0.86)$ & $-$ & $0.74,\ (0.50, 0.83)$ & $-$\\
\hline
\multirow{8}{*}{Power Consumption}
& KGD-D, 0.02 & $0.64,\ (0.51, 0.73)$ & $0.91   $ & $0.65,\ (0.50, 0.73)$ & $0.78   $\\
& KGD-D, 0.05 & $0.67,\ (0.54, 0.75)$ & $\bm{0.0019 }$ & $0.67,\ (0.55, 0.76)$ & $\bm{0.0004 }$\\
& KGD-D, 0.1  & $0.67,\ (0.55, 0.76)$ & $\bm{2.9e\text{-}5}$ & $0.67,\ (0.55, 0.75)$ & $\bm{1e\text{-}5  }$\\
& KGD-D, 0.2  & $0.65,\ (0.55, 0.74)$ & $0.057  $ & $0.64,\ (0.54, 0.74)$ & $0.21   $\\
& KGD-D, 0.5  & $0.61,\ (0.48, 0.68)$ & $1      $ & $0.59,\ (0.47, 0.67)$ & $1      $\\
\cline{2-6}
& KRR-CV      & $0.64,\ (0.50, 0.73)$ & $-$ & $0.63,\ (0.49, 0.73)$ & $-$\\
& KRR-MML     & $0.52,\ (0.42, 0.63)$ & $-$ & $0.51,\ (0.42, 0.61)$ & $-$\\
& KGF-CV      & $0.64,\ (0.48, 0.73)$ & $-$ & $0.63,\ (0.49, 0.72)$ & $-$\\
\hline
\multirow{8}{*}{Superconductors}
& KGD-D, 0.02 & $0.66,\ (0.54, 0.76)$ & $1      $ & $0.66,\ (0.54, 0.76)$ & $0.99   $\\
& KGD-D, 0.05 & $0.67,\ (0.57, 0.77)$ & $0.07   $ & $0.68,\ (0.57, 0.77)$ & $0.099  $\\
& KGD-D, 0.1  & $0.68,\ (0.57, 0.76)$ & $\bm{0.00033}$ & $0.68,\ (0.57, 0.76)$ & $\bm{0.0014 }$\\
& KGD-D, 0.2  & $0.67,\ (0.56, 0.74)$ & $0.081  $ & $0.67,\ (0.56, 0.74)$ & $0.25   $\\
& KGD-D, 0.5  & $0.61,\ (0.52, 0.70)$ & $1      $ & $0.60,\ (0.52, 0.69)$ & $1      $\\
\cline{2-6}
& KRR-CV      & $0.66,\ (0.55, 0.75)$ & $-$ & $0.66,\ (0.55, 0.75)$ & $-$\\
& KRR-MML     & $0.57,\ (0.47, 0.64)$ & $-$ & $0.56,\ (0.46, 0.64)$ & $-$\\
& KGF-CV      & $0.65,\ (0.54, 0.75)$ & $-$ & $0.65,\ (0.53, 0.75)$ & $-$\\
\hline
\multirow{8}{*}{U.K.\ Temperature}
& KGD-D, 0.02 & $0.50,\ (0.38, 0.61)$ & $1      $ & $0.50,\ (0.39, 0.61)$ & $1      $\\
& KGD-D, 0.05 & $0.53,\ (0.42, 0.64)$ & $0.42   $ & $0.53,\ (0.43, 0.64)$ & $0.052  $\\
& KGD-D, 0.1  & $0.55,\ (0.45, 0.64)$ & $\bm{0.0006 }$ & $0.55,\ (0.45, 0.64)$ & $\bm{3.6e\text{-}6}$\\
& KGD-D, 0.2  & $0.55,\ (0.44, 0.64)$ & $\bm{0.01   }$ & $0.55,\ (0.43, 0.64)$ & $\bm{0.005  }$\\
& KGD-D, 0.5  & $0.49,\ (0.38, 0.56)$ & $1      $ & $0.48,\ (0.36, 0.55)$ & $1      $\\
\cline{2-6}
& KRR-CV      & $0.54,\ (0.42, 0.63)$ & $-$ & $0.53,\ (0.42, 0.63)$ & $-$\\
& KRR-MML     & $0.43,\ (0.35, 0.50)$ & $-$ & $0.39,\ (0.31, 0.47)$ & $-$\\
& KGF-CV      & $0.53,\ (0.41, 0.63)$ & $-$ & $0.52,\ (0.41, 0.64)$ & $-$\\
\bottomrule
\end{tabular}
\label{tab:dec_2.5}
\end{table}

\newpage
\
\newpage
\section{Proofs}
\label{sec:proofs}

\begin{proof}[Proof of Theorem \ref{thm:gen_bound}]~\\
We first note that
\begin{equation*}
\begin{aligned}
&\E\left(\left|\fh(\xsv,\X,t)-f(\xsv)\right|\right)\\
&=\E\left(\left|\fh(\xsv,\X,t)+\frac1n\sum_{i=1}^n\fh(\xvi,\X,t)-\frac1n\sum_{i=1}^n\fh(\xvi,\X,t)-f(\xsv)\right|\right)\\
&\leq\E\left(\left|\fh(\xsv,\X,t)-\frac1n\sum_{i=1}^n\fh(\xvi,\X,t)\right|\right)+\E\left(\left|f(\xsv)-\frac1n\sum_{i=1}^n\fh(\xvi,\X,t)\right|\right).
\end{aligned}
\end{equation*}
For the first term, we use 
\begin{equation*}
\frac{\partial}{\partial t}\left(\fh(\xsv,\X,t)-\frac1n\sum_{i=1}^n\fh(\xvi,\X,t)\right)=\left(\kv(\xsv,\X,t)-\frac1n\sum_{i=1}^n\kv(\xvi,\X,t)\right)^\top\left(\yv-\fhv(\X,t)\right)
\end{equation*}
and $\|\yv-\fhv(\X,t)\|_2\leq e^{-\int_0^t\smin(\tau)d\tau}\cdot \|\yv\|_2$ from Lemma \ref{thm:yf_bound}
to obtain
\begin{equation}
\label{eq:term1}
\begin{aligned}
&\E\left(\left|\fh(\xsv,\X,t)-\frac1n\sum_{i=1}^n\fh(\xvi,\X,t)\right|\right)
=\left|\fh(\xsv,\X,t)-\frac1n\sum_{i=1}^n\fh(\xvi,\X,t)\right|\\
&=\left|\int_0^t\frac{\partial}{\partial \tau}\left(\fh(\xsv,\X,\tau)-\frac1n\sum_{i=1}^n\fh(\xvi,\X,\tau)\right)d\tau\right|\\
&\leq\int_0^t\left|\frac{\partial}{\partial \tau}\left(\fh(\xsv,\X,\tau)-\frac1n\sum_{i=1}^n\partial\fh(\xvi,\X,\tau)\right)\right|d\tau\\
&=\int_0^t\left|\left(\kv(\xsv,\X,\tau)-\frac1n\sum_{i=1}^n\kv(\xvi,\X,\tau)\right)^\top\left(\yv-\fhv(\X,\tau)\right)\right|d\tau\\
&\leq\int_0^t\left\|\kv(\xsv,\X,\tau)-\frac1n\sum_{i=1}^n\kv(\xvi,\X,\tau)\right\|_2\cdot \left\|\yv-\fhv\left(\X,\tau\right)\right\|_2d\tau\\
&=\frac1{\sqrt n}\cdot\dktwobar(\xsv,\X,t)\cdot\int_0^t\left\|\yv-\fhv\left(\X,\tau\right)\right\|_2d\tau\\
&\leq\frac1{\sqrt n}\cdot\dktwobar(\xsv,\X,t)\cdot\int_0^t e^{-\int_0^\tau\smin(\X,\tau_1)d\tau_1}d\tau\cdot\|\yv\|_2\\
&=\frac1{\sqrt n}\cdot\dktwobar(\xsv,\X,t)\cdot\int_0^t e^{-\tau\cdot\sminbar(\X,\tau)}d\tau\cdot\|\yv\|_2.
\end{aligned}
\end{equation}
For the second term, let $\bm{r}(\X,t):=\yv-\fhv(\X,t)$. Then $\fh(\xvi,\X,t)=y_i-r(\xvi,\X,t)=f(\xvi)+\varepsilon_i-r(\xvi,\X,t)$. Again, using $\|\yv-\fhv(\X,\tau)\|_2\leq e^{-\int_0^t\smin(\X,\tau)d\tau}\cdot \|\yv\|_2$ from Lemma \ref{thm:yf_bound}, and the mean value theorem, we obtain
\begin{equation}
\label{eq:term2}
\begin{aligned}
&\E\left(\left|f(\xsv)-\frac1n\sum_{i=1}^n\fh(\xvi,\X,t)\right|\right)
=\E\left(\left|f(\xsv)-\frac1n\sum_{i=1}^n\left(f(\xvi)+\varepsilon_i-r(\xvi,\X,t)\right)\right|\right)\\
&\leq\E\left(\left|f(\xsv)-\frac1n\sum_{i=1}^nf(\xvi)\right|\right)+\E\left(\left|\frac1n\sum_{i=1}^n r(\xvi,\X,t)-\frac1n\sum_{i=1}^n\varepsilon_i\right|\right)\\
&\leq\E\left(\frac1n\sum_{i=1}\left|f(\xsv)-f(\xvi)\right|\right)+\E\left(\frac1n\sum_{i=1}^n \left|r(\xvi,\X,t)\right|\right)+\E\left(\frac1n\sum_{i=1}^n\left|\varepsilon_i\right|\right)\\
&=\frac1n\sum_{i=1}\left|f(\xsv)-f(\xvi)\right|+\frac1n\sum_{i=1}^n \left|r(\xvi,\X,t)\right|+\frac1n\sum_{i=1}^n\E\left|\varepsilon\right|\\
&\leq \frac1n\sum_{i=1}^n\left|\left(\frac{\partial f(\bm{\tilde{x}_i})}{\partial \bm{\tilde{x}_i}}\right)^\top\cdot(\xsv-\xvi)\right|+\frac1n\left\|\bm{r}(\X,t)\right\|_1+\E|\varepsilon|\\
&\leq \frac1n\sum_{i=1}^n\left|\left(\frac{\partial f(\bm{\tilde{x}_i})}{\partial \bm{\tilde{x}_i}}\right)^\top\cdot(\xsv-\xvi)\right|+\frac1{\sqrt n}\left\|\yv-\fhv(\X,t)\right\|_2+\E|\varepsilon|\\
&\leq \frac1n\sum_{i=1}^n\left|\left(\frac{\partial f(\bm{\tilde{x}_i})}{\partial \bm{\tilde{x}_i}}\right)^\top\cdot(\xsv-\xvi)\right|+\frac1{\sqrt n}\|\yv\|_2\cdot e^{-t\cdot\sminbar(\X,t)}+\E|\varepsilon|,
\end{aligned}
\end{equation}
where $\bm{\tilde{x}_i}=(1-c_i)\cdot\xsv-c_i\cdot\xvi$ for some $c_i\in[0,1]$.

Putting Equations \ref{eq:term1} and \ref{eq:term2} together, we obtain
\begin{equation*}
\begin{aligned}
&\E_{\varepsilon}\left(\left|\fh(\xsv,\X,t)-f(\xsv)\right|\right)\\
\leq&\frac1{\sqrt n}\cdot\dktwobar(\xsv,\X,t)\cdot\int_0^t e^{-\tau\cdot\sminbar(\X,\tau)}d\tau\cdot\|\yv\|_2+\frac1n\sum_{i=1}^n\left|\left(\frac{\partial f(\bm{\tilde{x}_i})}{\partial \bm{\tilde{x}_i}}\right)^\top\cdot(\xsv-\xvi)\right|\\
&+\frac1{\sqrt n}\|\yv\|_2\cdot e^{-t\cdot\sminbar(\X,t)}+\E|\varepsilon|\\
=&\left(\dktwobar(\xsv,\X,t)\cdot\int_0^t e^{-\tau\cdot\sminbar(\X,\tau)}d\tau + e^{-t\cdot\sminbar(\X,t)}\right)\cdot\frac{\|\yv\|_2}{\sqrt n}\\
&+\underbrace{\frac1n\sum_{i=1}^n\left|\left(\frac{\partial f(\bm{\tilde{x}_i})}{\partial \bm{\tilde{x}_i}}\right)^\top\cdot(\xsv-\xvi)\right| +\E|\varepsilon|}_{=:C}\\
=:&\left(\dktwobar(\xsv,\X,t)\cdot\int_0^t e^{-\tau\cdot\sminbar(\X,\tau)}d\tau + e^{-t\cdot\sminbar(\X,t)}\right)\cdot\frac{\|\yv\|_2}{\sqrt n}+C.
\end{aligned}
\end{equation*}
\end{proof}

\begin{lemma}~\\
\label{thm:yf_bound}
\begin{equation*}
\left\|\yv-\fhv(\X,t)\right\|_2\leq e^{-\int_0^t \smin(\X,\tau)d\tau}\cdot\|\yv\|_2.
\end{equation*}
\end{lemma}

\begin{proof}~\\
The proof is an adaptation of the proof of Theorem 8.2 by \citet{rugh1996linear}.

Let $\etahv(t)=\yv-\fhv(t)$. Then $\etahv(0)=\yv$ and according to Equation \ref{eq:kgd_diff_eq}, 
\begin{equation*}
\etahv'(t)=-\frac{\partial \fhv(t)}{\partial t}=-\K(t)\left(\yv-\fhv(t)\right)=-\K(t)\etahv(t).
\end{equation*}
Using the chain rule, we obtain
\begin{equation*}
\left(\|\etahv(t)\|_2^2\right)'=\left(\etahv(t)^\top\etahv(t)\right)'=2\etahv(t)^\top\etahv'(t)=-2\etahv(t)^\top\K(t)\etahv(t)\leq -2\cdot\smin(t)\cdot\|\etahv(t)\|_2^2,
\end{equation*}
where in the last inequality we have used that the Rayleigh quotient of a symmetric matrix, $\A$, is bounded by its eigenvalues: For all vectors $\vv$,
\begin{equation*}
\smin(\A)\leq\frac{\vv^\top\A\vv}{\vv^\top\vv}\leq\smax(\A)\iff\smin(\A)\cdot\|\vv\|_2^2\leq\vv^\top\A\vv\leq\smax(\A)\cdot\|\vv\|_2^2.
\end{equation*}
Using Grönwall's inequality on
\begin{equation*}
\left(\|\etahv(t)\|_2^2\right)'\leq -2\smin(t)\cdot\|\etahv(t)\|_2^2,
\end{equation*}
we obtain

\begin{equation*}
\begin{aligned}
\|\etahv(t)\|_2^2&\leq e^{-2\int_0^t\smin(\tau)d\tau}\cdot\|\etahv(0)\|_2^2\iff\|\etahv(t)\|_2\leq  e^{-\int_0^t\smin(\tau)d\tau}\cdot\|\etahv(0)\|_2\\
&=\underbrace{e^{-\int_0^t\smin(\tau)d\tau}}_{\leq 1}\cdot \|\yv\|_2\leq\|\yv\|_2.
\end{aligned}
\end{equation*}
\end{proof}

\begin{proof}[Proof of Proposition \ref{thm:dktwo_bd}]~\\
The first inequality is just a simple consequence of Jensen's inequality: $(\E(\|\vv\|_2))^2\leq\E(\|\vv\|_2^2))$.

For any random vector $\vv$, with $\E(\vv)=\muv$ and $\Cov(\vv)=\Ss$,
\begin{equation}
\label{eq:exp_norm}
\begin{aligned}
\E(\|\vv\|_2^2)
&= \E\left(\|\muv+(\vv-\muv)\|_2^2\right)
= \|\bm{\mu}\|_2^2+2\bm{\mu}^\top\underbrace{\E(\vv-\muv)}_{=\nv}+\E\left((\vv-\muv)^\top(\vv-\muv)\right)\\
&= \|\bm{\mu}\|_2^2+\E\left(\Tr\left((\vv-\muv)^\top(\vv-\muv)\right)\right)
= \|\bm{\mu}\|_2^2+\Tr\left(\E\left((\vv-\muv)(\vv-\muv)^\top\right)\right)\\
&= \|\bm{\mu}\|_2^2+\Tr(\bm{\Sigma}).
\end{aligned}
\end{equation}

Let $\dkv(\xsv,\X):=\kv(\xsv,\X)-\frac1n\sum_{i=1}^n\kv(\xvi,\X)$.
To calculate $\E(\dktwo(\xsv,\X)^2)=\E(n\cdot\|\dkv(\xsv,\X)\|_2^2)$, we need to calculate $\E(\dkv(\xsv,\X))$ and the diagonal elements of $\Cov(\dkv(\xsv,\X))$. We do this seperately for each element, $i\in\{1,\dots n\}$:

\begin{equation*}
\begin{aligned}
\E(\dkv(\xsv,\X))_i&=\E\left(k(\xsv,\xvi)-\frac1n\sum_{j=1}^nk(\xvi,\xvj)\right)\\
&=\E\left(k(\xsv,\xvi)\right)-\frac1n\left(\E(k(\xvi,\xvi))+\sum_{\substack{j=1\\j\neq i}}^n\E\left(k(\xvi,\xvj)\right)\right)\\
&=\frac nn\mu_k-\frac1n\left(k(0)+(n-1)\mu_k\right)
=\frac1n\left(n\mu_k-k(0)-(n-1)\mu_k\right)\\
&=\frac{\mu_k-k(0)}n,
\end{aligned}
\end{equation*}
and
\begin{equation*}
\begin{aligned}
\Cov(\dkv(\xsv,\X))_{ii}=&\Var\left(\frac nnk(\xsv,\xvi)-\frac1n\sum_{j=1}^nk(\xvi,\xvj)\right)\\
=&\frac1{n^2}\Var\left(nk(\xsv,\xvi)-k(\xvi,\xvi)-\sum_{\substack{j=1\\j\neq i}}^nk(\xvi,\xvj)\right)\\
=&\frac1{n^2}\left(n^2\varsigma_k^2+(n-1)\varsigma_k^2-2n(n-1)c_k+\left((n-1)^2-(n-1)\right)c_k\right)\\
=&\frac{(n^2+n-1)(\varsigma_k^2-c_k)+c_k}{n^2}.
\end{aligned}
\end{equation*}

Putting the expressions for $\E(\dkv(\xsv,\X))_i$ and $\Cov(\dkv(\xsv,\X))_{ii}$ into Equation \ref{eq:exp_norm}, we obtain
\begin{equation*}
\begin{aligned}
\E(\dktwo(\xsv,\X)^2)&=
\E(n\cdot\|\dkv(\xsv,\X)\|_2^2)\\
&=n\cdot\left(\sum_{i=1}^n (\E(\dkv(\xsv,\X))_i)^2 + \sum_{i=1}^n \E(\dkv(\xsv,\X))_{ii}\right)\\
&=n\left(n\cdot\frac{(\mu_k-k(0))^2}{n^2}+n\cdot\frac{(n^2+n-1)(\varsigma_k^2-c_k)+c_k}{n^2}\right)\\
&=(\mu_k-k(0))^2+(n^2+n-1)\cdot(\varsigma_k^2-c_k)+c_k\\
&=(\mu_k-k(0))^2+(n^2+n-1)\cdot(\varsigma_k^2-c_k)-(\varsigma_k^2-c_k)+\varsigma_k^2\\
&=(\mu_k-k(0))^2+(n^2+n-2)\cdot(\varsigma_k^2-c_k)+\varsigma_k^2.
\end{aligned}
\end{equation*}
Since  $n^2+n-2\geq 0$ for $n\geq 1$, we just need to make sure that $\varsigma^2-c_k\geq 0$:
\begin{equation*}
\begin{aligned}
\varsigma^2-c_k&=\Var\left(k(\xv,\xpv)\right)-\Cov\left(k(\xv,\xpv),k(\xv,\bm{x''})\right)\\
&=\frac12\left(\Var\left(k(\xv,\xpv)\right)+\underbrace{\Var\left(k(\xv,\bm{x''})\right)}_{=\Var\left(k(\xv,\xpv)\right)}-2\Cov\left(k(\xv,\xpv),k(\xv,\bm{x''})\right)\right)\\
&=\frac12\Var\left(k(\xv,\xpv)-k(\xv,\bm{x''})\right)\geq 0.
\end{aligned}
\end{equation*}
\end{proof}

\begin{proof}[Proof of Proposition \ref{thm:dr2dt}]~\\
Using the chain rule and $\partial_t \fhv(t)=\K(t)(\yv-\fhv(t))$, from Equation \ref{eq:kgd_diff_eq}, we obtain
\begin{equation*}
\begin{aligned}
\frac{\partial \Rttr(t)}{\partial t}=&\left(\frac{\partial \Rttr(t)}{\partial \fhv(t)}\right)^\top\left(\frac{\partial \fhv(t)}{\partial t}\right)=\frac{2\left(\yv-\fhv(t)\right)^\top}{\left\|\yv\right\|_2^2}\K(t)\left(\yv-\fhv(t)\right)\\
=&2\cdot\frac{\left\|\yv-\fhv(t)\right\|^2_{\K(t)}}{\left\|\yv\right\|_2^2}\geq 0,
\end{aligned}
\end{equation*}
which proves Equation \ref{eq:dr2dt1}.

For a constant $\K$,
\begin{equation*}
\begin{aligned}
\frac{\partial^2 \Rttr(t)}{\partial t^2}=&\left(\frac{\partial \frac{\partial \Rttr(t)}{\partial t}}{\partial \fhv(t)}\right)^\top\left(\frac{\partial \fhv(t)}{\partial t}\right)=\frac{-4\left(\yv-\fhv(t)\right)^\top\K}{\left\|\yv\right\|_2^2} \cdot \K\left(\yv-\fhv(t)\right)\\
=&-4\cdot\frac{\left\|\yv-\fhv(t)\right\|^2_{\K^2}}{\left\|\yv\right\|_2^2}\leq 0,
\end{aligned}
\end{equation*}
which proves Equation \ref{eq:dr2dt3}.
\end{proof}

\begin{proof}[Proof of Proposition \ref{thm:sigma_compl}]~\\
Let $\bm{1}\in\R^n$ denote of vector of only ones. We first use the matrix inversion lemma,
\begin{equation*}
\D^{-1}\C(\A-\B\D^{-1}\C)^{-1}=(\D-\C\A^{-1}\B)^{-1}\C\A^{-1},
\end{equation*}
for $\A=\lambda\I$, $\B=\bm{1}$, $\C=\bm{1}^\top$ and $\D=-\I$, to show that
\begin{equation*}
\bm{1}^\top(\bm{1}\bm{1}^\top+\lambda\I)^{-1}
=(\I+\bm{1}^\top(1/\lambda\I)\bm{1})^{-1}\bm{1}^\top(1/\lambda\I)
=1/\lambda(\I+1/\lambda\bm{1}^\top\bm{1})^{-1}\bm{1}^\top
=\frac1{n+\lambda}\cdot\bm{1}^\top.
\end{equation*}

Now, since $\K(\sigma=\infty)=\bm{1}\bm{1}^\top$,
\begin{equation*}
\begin{aligned}
&\df\left(\fh(\xsv,\X,\lambda,\sigma=\infty)\right)
=\Tr\left(\K(\sigma=\infty)\left(\K(\sigma=\infty)+\lambda\I\right)^{-1}\right)\\
&=\Tr\left(\bm{1}\bm{1}^\top\left(\bm{1}\bm{1}^\top+\lambda\I\right)^{-1}\right)\
=\Tr\left(\frac1{n+\lambda}\bm{1}\bm{1}^\top\right)
=\frac n{n+\lambda}=\frac1{1+\lambda/n},
\end{aligned}
\end{equation*}
and since $\kv(\xsv,\X,\sigma=\infty)=\bm{1}$,
\begin{equation*}
\begin{aligned}
\fh(\xsv,\X,\lambda,\sigma=\infty)
&=\kv(\xsv,\sigma=\infty)^\top\left(\K(\sigma=\infty)+\lambda\I\right)^{-1}\yv
=\bm{1}^\top(\bm{1}\bm{1}^\top+\lambda\I)^{-1}\yv\\
&=\frac1{n+\lambda}\cdot\sum_{i=1}^ny_i
=\frac1{1+\lambda/n}\cdot\frac1n\sum_{i=1}^ny_i.
\end{aligned}
\end{equation*}

Since $\K(\sigma=0)=\I$,
\begin{equation*}
\begin{aligned}
\df\left(\fh(\xsv,\X,\lambda,\sigma=0)\right)
&=\Tr\left(\K(\sigma=0)\left(\K(\sigma=0)+\lambda\I\right)^{-1}\right)
=\Tr\left(\I\left(\I+\lambda\I\right)^{-1}\right)\\
&=\frac n{1+\lambda}.
\end{aligned}
\end{equation*}
Furthermore, since
$\kv(\xsv,\X,\sigma=0)=\begin{cases}
\nv\text{ if } \xsv \notin \X\\
\bm{e_i}\text{ if } \xsv = \xvi\in \X,
\end{cases}$
where $\bm{e_i}$ is a vector of only zeros, except element $i$ which is one, if $\xsv \notin \X$,
\begin{equation*}
\begin{aligned}
\fh(\xsv,\X,\lambda,\sigma=0)
&=\kv(\xsv,\X,\sigma=0)^\top\left(\K(\sigma=0)+\lambda\I\right)^{-1}\yv
=\bm{0}^\top(\I+\lambda\I)^{-1}\yv
=0,
\end{aligned}
\end{equation*}
while if $\xsv =\xvi \in \X$,
\begin{equation*}
\begin{aligned}
\fh(\xsv,\X,\lambda,\sigma=0)
&=\kv(\xsv,\X,\sigma=0)^\top\left(\K(\sigma=0)+\lambda\I\right)^{-1}\yv
=\bm{e_i}^\top(\I+\lambda\I)^{-1}\yv\\
&=\frac1{1+\lambda}\bm{e_i}^\top\yv
=\frac{1}{1+\lambda}\cdot y_i.
\end{aligned}
\end{equation*}
\end{proof}

\begin{proof}[Proof of Proposition \ref{thm:der_bound_lbda}]~\\

\begin{equation*}
\begin{aligned}
\left\|\frac{\partial \fh(\xsv,\X,\lambda)}{\partial \xsv}\right\|_2
&=\left\|\frac{\partial}{\partial \xsv}\left(\kv(\xsv,\X)^\top(\K+\lambda\I)^{-1}\yv\right)\right\|_2\\
&=\left\|\left(\frac{\partial \kv(\xsv,\X)}{\partial \xsv}\right)^\top\cdot(\K+\lambda\I)^{-1}\yv\right\|_2\\
&\leq\left\|\frac{\partial \kv(\xsv,\X)}{\partial \xsv}\right\|_2\cdot\left\|(\K+\lambda\I)^{-1}\right\|_2\cdot\left\|\yv\right\|_2.
\end{aligned}
\end{equation*}

We bound the first two factors separately:
\begin{equation*}
\begin{aligned}
&\left\|\frac{\partial \kv(\xsv,\X)}{\partial \xsv}\right\|_2
=\left\|\frac{\partial}{\partial \xsv}\left(\left[k\left(\frac{\|\xsv-\bm{x_1}\|_2}{\sigma}\right),\ k\left(\frac{\|\xsv-\bm{x_2}\|_2}{\sigma}\right)\dots,\ k\left(\frac{\|\xsv-\bm{x_n}\|_2}{\sigma}\right)\right]\right)\right\|_2\\
&=\left\|\left[k'\left(\frac{\|\xsv-\bm{x_1}\|_2}{\sigma}\right)\cdot\frac1{\sigma}\cdot \frac{\xsv-\bm{x_1}}{\|\xsv-\bm{x_1}\|_2}\dots,\ k'\left(\frac{\|\xsv-\bm{x_n}\|_2}{\sigma}\right)\cdot\frac1{\sigma}\cdot \frac{\xsv-\bm{x_n}}{\|\xsv-\bm{x_n}\|_2}\right]\right\|_2\\
&\leq \frac{k'_{\max}}{\sigma}\cdot \left\|\left[\frac{\xsv-\bm{x_1}}{\left\|\xsv-\bm{x_1}\right\|_2}\dots,\ \frac{\xsv-\bm{x_n}}{\left\|\xsv-\bm{x_n}\right\|_2}\right]\right\|_2
\leq \frac{k'_{\max}}{\sigma}\cdot \left\|\left[\frac{\xsv-\bm{x_1}}{\left\|\xsv-\bm{x_1}\right\|_2}\dots,\ \frac{\xsv-\bm{x_n}}{\left\|\xsv-\bm{x_n}\right\|_2}\right]\right\|_F\\
&= k'_{\max}\cdot\frac1{\sigma}\cdot \sqrt{n},
\end{aligned}
\end{equation*}
where in the last inequality we used $\|\A\|_2\leq\|\A\|_F$, and in the last equality we used that each column in the $n\times p$ matrix has Euclidean norm 1.
\end{proof}

\end{document}